\documentclass[lettersize,journal]{IEEEtran}
\usepackage{amsmath,amsfonts}
\usepackage{algorithmic}
\usepackage{algorithm}
\usepackage{array}
\usepackage[caption=false,font=normalsize,labelfont=sf,textfont=sf]{subfig}
\usepackage{textcomp}
\usepackage{stfloats}
\usepackage{url}
\usepackage{verbatim}
\usepackage{graphicx}
\usepackage{cite}
\usepackage{multirow}
\usepackage[table,xcdraw]{xcolor}
\usepackage{hyperref}
\usepackage{orcidlink} 
\usepackage{orcidlink} 
\begin{document}

\title{Query, Align, and Distill: Navigation-Aware Cross-Modal Interaction for Efficient Vision-and-Language Navigation}

\author{
Zhihao~Chen\textsuperscript{*}\,\textsuperscript{\orcidlink{0009-0000-6806-1326}},
Yiyuan~Ge\textsuperscript{*}\,\textsuperscript{\orcidlink{0009-0006-5442-1865}},
Ziyang~Wang\,\textsuperscript{\orcidlink{0000-0003-1605-0873}},
~\IEEEmembership{Senior~Member,~IEEE},
Pu~Cao\,\textsuperscript{\orcidlink{0000-0003-4717-1539}},
~Lu~Yang\,\textsuperscript{\orcidlink{0000-0001-6745-2261}},
~\IEEEmembership{Member,~IEEE}
\thanks{Zhihao Chen, Pu Cao, and Lu Yang are with the School of
Intelligent Engineering and Automation, Beijing University of Posts
and Telecommunications, Beijing, China.

Yiyuan Ge is with the School of Electronic and Information Engineering,
South China University of Technology, Guangzhou, China.

Ziyang Wang is with the School of Computer Science and Digital
Technologies, Aston University, Birmingham, U.K.\\
\textsuperscript{*} Equal Contribution \\
Corresponding author: Lu Yang (soeaver@bupt.edu.cn).}
}


\markboth{Journal of \LaTeX\ Class Files,~Vol.~14, No.~8, August~2021}%
{Shell \MakeLowercase{\textit{et al.}}: A Sample Article Using IEEEtran.cls for IEEE Journals}


\maketitle

\begin{abstract}
Recent large-scale Vision-and-Language Navigation (VLN) models deliver strong accuracy but remain costly to deployment due to heavy parameters and computation. 
We tackle efficient VLN in two steps. First, we build a high-performing teacher that makes navigation evidence selection explicit and compressible. 
Concretely, the teacher introduces a small set of learnable query slots to extract global and local action-sufficient navigable evidence from panoramic observations via a Navigable Query Generator, and progressively grounds these evidence tokens to the instruction with an Instruction–Query Aligner for policy prediction. 
Second, leveraging this explicit query bottleneck as a distillation interface, we train a compact student by transferring both where to attend and what to do: we distill the teacher’s global/local navigable queries with a navigation-aware token-adaptive objective, and further match action distributions during fine-tuning. 
Experiments on standard VLN benchmarks demonstrate that our student nearly matches the teacher’s navigation performance while reducing the number of parameters by 93.65\% compared to the teacher.
\end{abstract}


\section{Introduction}
Vision-and-Language Navigation (VLN)~\cite{he2025navcomposer, zhan2024enhancing, zhong2026spatial, zhang2020language, yuan2026spenav} is the task in which an embodied agent follows natural-language instructions to navigate through observed environments.
As a fundamental embodied AI capability, it enables agents to execute human instructions in real environments, e.g., household assistance and navigation support.
However, modern VLN systems often rely on complex architectures and heavy cross-modal interaction layers to achieve strong navigation accuracy, leading to substantial compute and memory overhead that hinders deployment on resource-limited edge platforms.

To tackle the above efficiency bottleneck, recent VLN research introduces knowledge distillation (KD), training a compact student under the guidance of a teacher model to reduce computation while retaining strong navigation performance in resource-limited settings. 
Existing VLN distillation can be organized into two paradigms: (i) representation distillation~\cite{2023Knowledge,elnoor2025vi} during the pre-training stage, which transfers fine-grained cross-modal feature/representation alignment (e.g., embeddings/attentions/hidden states), and (ii) decision distillation~\cite{wang2024magic, zhu2025minivln} during the fine-tuning stage, which transfers navigation-specific action determination signals (e.g., fused logits/policy).
However, we observe two practical obstacles in VLN. 
First, navigation errors compound over time, making KD strongly dependent on teacher quality, and a suboptimal teacher sets a low performance ceiling for the student.
% Second, most existing KD methods focus on distilling what to predict (e.g., features or actions), but fail to transfer how the teacher selects visual evidence for decision-making, leaving this critical process implicit.
\textcolor{black}{Second, existing VLN distillation methods transfer backbone representations, attention patterns, or action distributions, but they do not explicitly separate local cues for immediate action selection from global cues for long-term route planning.}

Motivated by this, we address the above challenges in two steps. First, we train a high-performing teacher with a navigation-aware cross-modal model that makes evidence selection explicit. 
Concretely, at each time step, a Navigable Query Generator introduces a small set of learnable query slots to selectively extract action-relevant evidence from panoramic observations, yielding global navigable queries that summarize long-horizon context and local navigable queries that focus on immediate, action-feasible directions. 
An Instruction–Query Aligner then progressively grounds and fuses instruction semantics with these queries to form a compact cross-modal state for action prediction.

\begin{figure}[!t]
\centering
\includegraphics[width=\linewidth]{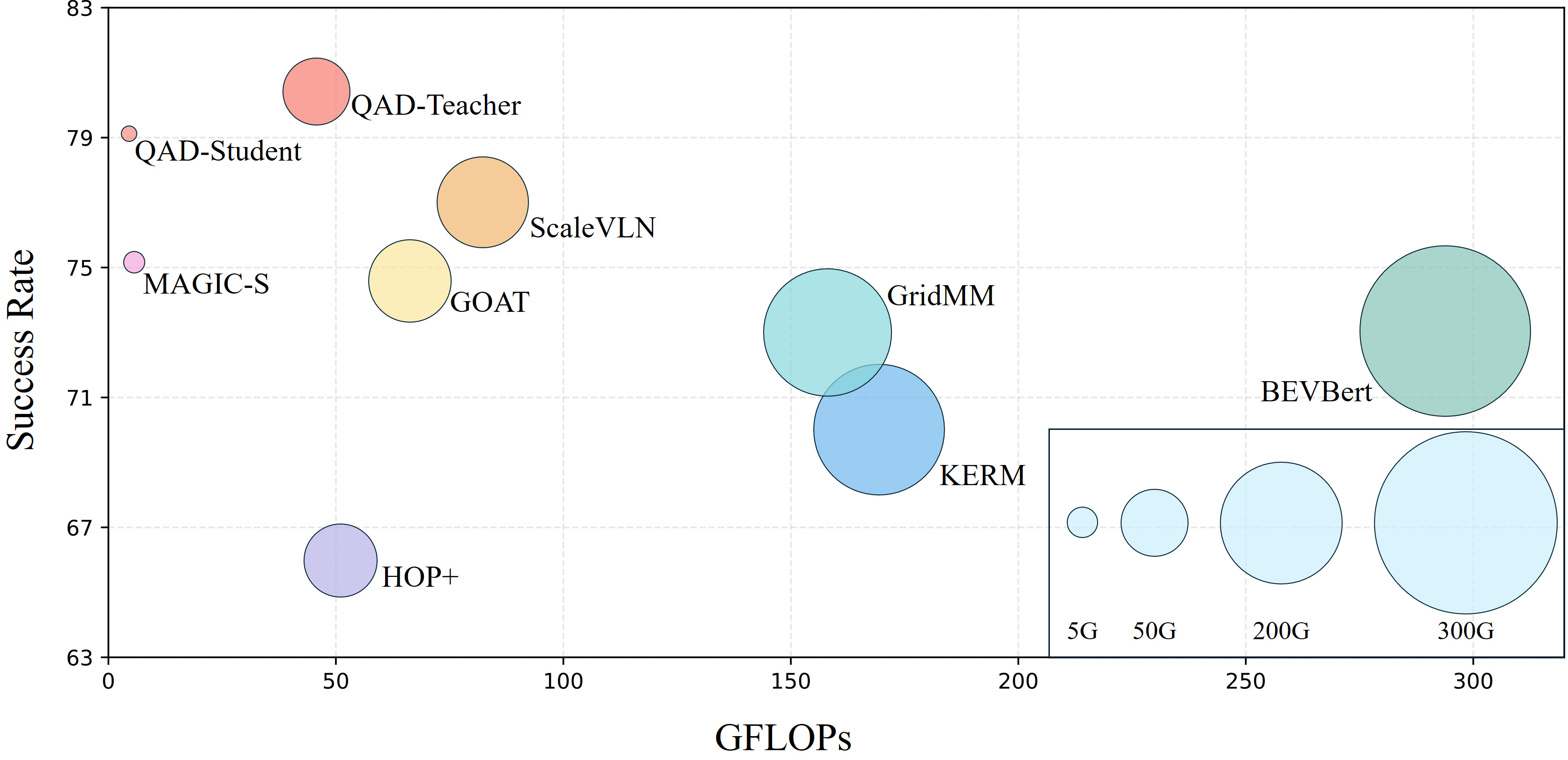}
\caption{\textcolor{black}{
Accuracy–efficiency trade-off on R2R (test unseen). 
We plot success rate (SR) against model computation (GFLOPs). 
Our compact student achieves near-teacher performance with substantially lower computation.
Bubble size represents GFLOPs, with the 5G--300G reference bubbles indicating the scale.
}
}
\label{fig:fig1}
\end{figure}

% Second, with this explicit evidence-selection bottleneck as a distillation interface, we train a compact student via Navigation-aware Token-adaptive Distillation. 
% Specifically, we distill the teacher’s global and local navigable queries in both pre-training and fine-tuning, and further transfer action distributions via a policy KL objective during fine-tuning. 
\textcolor{black}{Second, we reuse the same local/global query representations as the primary teacher–student transfer targets. 
Navigation-aware Token-adaptive Distillation matches teacher and student representations at the individual query-slot level and assigns larger weights to poorly aligned pairs, while policy KL is used as complementary behavior-level supervision during fine-tuning. }
By inheriting the teacher’s evidence-selection capability through query-level distillation, QAD achieves a favorable accuracy–efficiency trade-off: as shown in \textcolor{black}{Fig.}~\ref{fig:fig1}, the student attains higher success rates at lower GFLOPs than prior methods while substantially reducing model size and computation. In summary, our contributions are as follows:

We present a teacher–student framework called QAD that explicitly exposes evidence selection via a compact set of navigable queries, making cross-modal reasoning more amenable to compression for efficient VLN. QAD instantiates a Navigable Query Generator to extract local and global navigable evidence tokens, and an Instruction–Query Aligner to progressively ground and fuse instruction semantics with these tokens for action prediction. We further propose Navigation-aware Token-adaptive Distillation (with policy distillation) to train a compact student that nearly matches the teacher while cutting parameters by 93.65\% (150.78M→9.56M) and computational cost from 45.56 to 1.92 GFLOPs.

\section{Related Work}
\subsection{Vision-and-Language Navigation}
Vision-and-Language Navigation (VLN) is the task of following natural-language instructions to navigate in a visual environment towards a target.
Its development can be summarized in four stages: (1) Seq2Seq instruction-following agents, which established the canonical instruction-to-action paradigm via imitation and reinforcement learning in photorealistic simulators. In this line, R2R~\cite{anderson2018vision} on Matterport3D set standard evaluation protocols, while data augmentation and pragmatic inference (e.g., Speaker-Follower~\cite{fried2018speaker}), together with improved grounding and generalization (e.g., RCM~\cite{wang2019reinforced} with SIL), alleviated limited supervision and narrowed the seen–unseen gap. 
Zhang et al.~\cite{zhang2020language} strengthen instruction--observation correspondence through historical and mutual co-grounding and alleviate the training--inference discrepancy via alternate adversarial learning.
More recently, NavComposer~\cite{he2025navcomposer} addresses data scarcity by decomposing navigation trajectories into action, scene, and object entities and recomposing them into high-quality instructions.
(2) Transformer-based grounding and pretraining shifted focus to stronger cross-modal token interactions and transfer learning, including VLN-specific pretraining~\cite{hao2020towards}, history-aware transformer designs~\cite{hong2021vln}, in-domain pretraining~\cite{Guhur_2021_ICCV}, and auxiliary self-supervised reasoning signals~\cite{Zhu_2020_CVPR} to improve robustness and generalization. 
Zhan et al.~\cite{zhan2024enhancing} inject room-type scene knowledge into the navigator through multimodal learnable prompt pools to improve cross-modal grounding.
(3) History-/map-aware planning introduced explicit memory and structured spatial representations for long-horizon and goal-driven navigation, such as hierarchical history transformers~\cite{chen2021history}, dual-scale topological-map reasoning~\cite{chen2022think}, and dynamically growing grid memories~\cite{wang2023gridmm}. 
Zhong et al.~\cite{zhong2026spatial} construct a multi-layer spatial map spanning waypoints, objects, rooms, and floors and use 3D Gaussian splatting to obtain viewpoint-robust features.
SPENav~\cite{yuan2026spenav} combines open-vocabulary perception, dynamic object filtering, and spatial--instructional attention to build task-oriented selective memory.
(4) Foundation-model-driven VLN leverages LLM/VLM priors for explicit reasoning and task unification, ranging from LLM-centric decision making with interpretable traces~\cite{Zhou_Hong_Wu_2024} to tighter VLM–LLM alignment~\cite{zhou2024navgpt} and generalist embodied navigation training across heterogeneous datasets, as well as fast/slow hierarchical reasoning systems that activate expensive VLM reasoning only when needed~\cite{zhou2025fsr}.

\begin{figure*}[t]
\centering
\includegraphics[width=\linewidth]{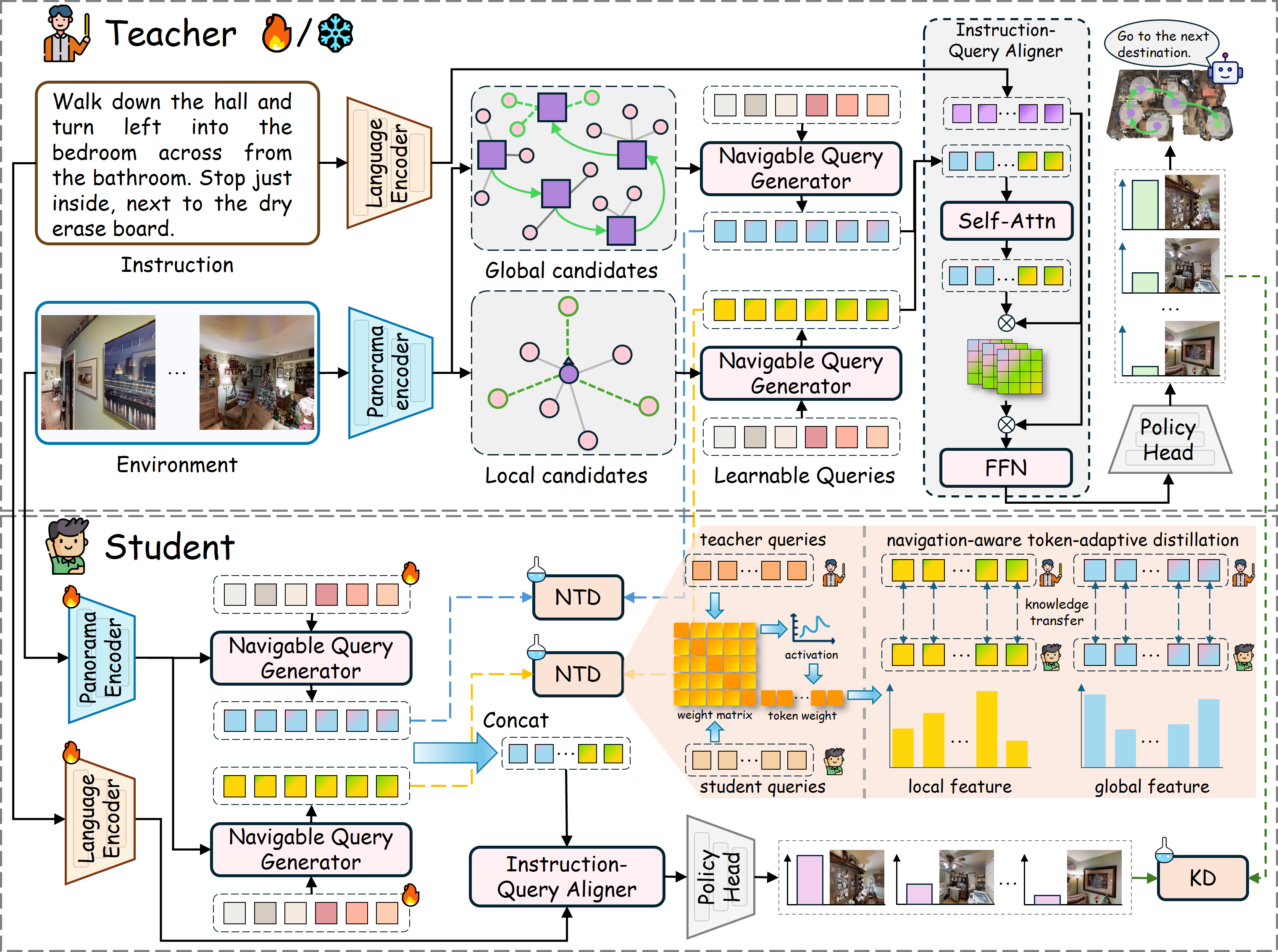}
\caption{\textcolor{black}{Overview of Query, Align, and Distill. 
Following the global-planning and local-acting paradigm in Vision-and-Language Navigation, history-based global candidates support long-range planning, while current-view local candidates support fine-grained action selection. 
The Navigable Query Generator extracts action-sufficient global and local queries, which the Instruction--Query Aligner grounds with the instruction for policy prediction. 
The shallower student is trained by Navigation-aware Token-adaptive Distillation and policy distribution distillation during fine-tuning. 
Blue and orange dashed lines denote global and local query distillation, respectively. 
Flame and snowflake icons mark trainable and frozen modules, respectively.}}
\label{fig:framework}
\end{figure*}

\subsection{Cross-Modal Interaction in VLN}
At each step in VLN, an agent must ground instruction semantics in panoramic observations to decide the next action, making cross-modal interaction central to the task.
Most methods instantiate this grounding via vision--language attention that repeatedly exchanges information between instruction tokens and visual tokens.
Some architectures incorporate time- or history-aware fusion; for example, VLN-BERT~\cite{hong2021vln} introduces a recurrent cross-modal state, and HAMT~\cite{chen2021history} employs a history-aware multimodal transformer over long-horizon observations.

Map-aware interaction further couples grounding with planning by performing multimodal reasoning on explicit spatial structures, e.g., CM2~\cite{Georgakis_2022_CVPR} grounds language on learned egocentric top-down semantic maps, and DUET~\cite{chen2022think} combines fine-scale language grounding with coarse-scale reasoning on an online topological map via dual-scale graph transformers.

\textcolor{black}{
Query- and slot-based mechanisms provide another route to compact visual evidence.
Slot Attention~\cite{locatello2020object} iteratively groups perceptual features into exchangeable object-centric slots, while VLN variants such as Local Slot Attention~\cite{zhuang2022local} and GeoVLN~\cite{huo2023geovln} mainly aggregate object-level or geometry-aware local visual cues.
Related selective-attention designs such as SKNet~\cite{li2019selective} instead adaptively fuse convolutional branches with different receptive fields.
}

Pretraining objectives also strengthen navigation-oriented alignment, e.g., AirBERT's~\cite{Guhur_2021_ICCV} in-domain pretraining and CSAP's~\cite{10.1145/3503161.3548283} cross-modal semantic alignment pretraining.
Beyond post-fusion attention, some studies emphasize improving alignment quality itself: DELAN~\cite{du2024delan} performs dual-level \emph{pre-fusion} alignment (instruction--history and landmark--observation) via cross-modal contrastive learning to better calibrate modality spaces before fusion, while syntactic-structure-aware VLN models inject dependency/phrase structures to localize action-relevant sub-instructions, suggesting that phrase-level organization can improve grounding and generalization in unseen environments~\cite{li2021improving}.

\textcolor{black}{
Overall, QAD builds on DUET's dual-scale navigation substrate but differs from both dense token interaction and prior slot-based visual aggregation.
Rather than learning exchangeable object-centric slots or kernel-selection weights, QAD extracts branch-specific navigation-evidence queries from one-step navigable candidates and the global topological graph before instruction grounding.
The local and global query sets encode immediate actionability and long-horizon topological progress, respectively, and are further reused as an explicit teacher--student distillation interface.
}

\subsection{Knowledge Distillation in VLN}
Knowledge distillation (KD) has recently emerged as a practical route for achieving efficiency in VLN by transferring navigation-relevant capabilities from a large teacher to a lightweight student.
Early attempts distill VLN pretraining knowledge into a smaller vision--language backbone to reduce parameters and inference cost while retaining most performance~\cite{2023Knowledge}.
MAGIC~\cite{wang2024magic} jointly transfers attention patterns, intermediate features, and action logits, so the student can inherit both cross-modal alignment behavior and decision capability from the teacher.
\textcolor{black}{In comparison, NTD takes the local/global navigable-query embeddings that explicitly encode the teacher's selected navigation evidence as its primary transfer targets, and adjusts supervision at the individual query-slot level.}
MiniVLN~\cite{zhu2025minivln} adopts progressive two-stage distillation to capture fine-grained knowledge during pretraining and navigation-specific knowledge during fine-tuning, producing a compact student with near-teacher performance.
Relatedly, distillation has also been explored at the level of cross-modal interaction mechanisms.
Vi-LAD~\cite{elnoor2025vi} distills vision--language attention for socially-aware robot navigation in dynamic environments, highlighting that transferring attention-based evidence selection can be beneficial beyond logits.
In the broader vision--language understanding literature, distilled dual-encoder models transfer cross-modal attention distributions from a fusion teacher to an efficient dual-encoder student, strengthening the connection between interaction distillation and efficient inference~\cite{wang2022distilled}.
Overall, prior KD methods mainly transfer backbone representations, attention behavior, or output decisions.
QAD instead distills a structured navigation-evidence interface through slot-adaptive local/global query matching, while using policy KL only as complementary behavior-level supervision.

\section{Method}
\subsection{Problem Formulation}
Vision-and-Language Navigation (VLN) trains an embodied agent to follow a natural-language instruction and navigate in an unseen environment.
We model the environment as a connectivity graph $G=(V,E)$, where each node $v \in V$ represents a viewpoint, and each edge $(v_i, v_j) \in E$ indicates navigability between two viewpoints.
Given an instruction $I = \{w_i\}^L_{i=1}$ with $L$ words and an initial viewpoint $v_1$, the agent interacts with the environment over a sequence of time steps $t=1,\dots,T$. 
At step $t$, the agent is located at viewpoint $v_t$, receives a panoramic observation $O_t$, and is provided with a set of navigable candidates $\mathcal{N}(v_t)=\{u|(v_t,u) \in \mathcal{E} \}$. 
Conditioned on the instruction, the current observation, and the navigation history, the agent selects an action from a discrete set $\mathcal{A}_t = \mathcal{N}(v_t) \cup \{STOP\}$, executing $a_t \in \mathcal{N}(v_t)$ moves to next viewpoint $v_{t+1}$, while $STOP$ terminates current navigation episode. 
The objective is to learn a policy $\pi_{\theta}(a_t|I,O_{\le t}, a_{<t})$ that generates a trajectory $\tau = (v_1, a_1, \dots, v_T, a_T)$ following the instruction, and success is achieved when the stop location falls within a fixed threshold of the ground-truth goal. 
For goal-oriented benchmarks (e.g., REVERIE~\cite{qi2020reverie}, SOON~\cite{zhu2021soon}), the agent additionally needs to localize the target object at the final viewpoint.

\subsection{Overview}
As illustrated in \textcolor{black}{Fig.}~\ref{fig:framework}, we propose a teacher-student framework for efficient VLN.
Given the panoramic observation $O_t$ and instruction $I$, the teacher first extracts action-relevant visual evidence via a \textbf{Navigable Query Generator}, producing a set of global and local navigable queries $\mathbf{Q}^g_t$ and $\mathbf{Q}^l_t$.
These queries are then fused with instruction representations by an \textbf{Instruction-Query Aligner} to obtain a cross-modal state $\mathbf{z}_t$, which is finally fed into a policy head to predict the next action distribution $\pi(a_t|\cdot)$.
To achieve a compact yet accurate student, we keep the same overall architecture but reduce the depth of the language and panorama encoder.
We distill the teacher’s navigation intent to the student through Navigation-aware Token-adaptive Distillation (NTD) in both pre-training and fine-tuning stages, and we additionally distill the task-specific policy distribution via a KL divergence loss during fine-tuning.

\subsection{Navigable Query Generator (Query)}
Similar to DUET\cite{chen2022think} and its follow-ups \cite{10801563}, we maintain a local branch that encodes fine-grained features $v^L_t\in \mathbb{R}^{N_L \times d}$ from the current viewpoint for one-step candidates $\mathcal{N}(v_t)$, alongside a global branch that maintains a topological graph $G_t$ with coarse-grained node features $v^G_t \in \mathbb{R}^{N_G \times d}$ over all visited viewpoints. 
Building on this dual-branch structure, we introduce a Navigable Query Generator that converts $v^L_t$ and $v^G_t$ into compact, action-sufficient local and global navigable queries. These queries serve as explicit evidence tokens for subsequent alignment and decision making.

Formally, we parameterize two sets of learnable query slots,
$\mathbf{Q}^{L}_{0}\in\mathbb{R}^{K_L\times d}$ and $\mathbf{Q}^{G}_{0}\in\mathbb{R}^{K_G\times d}$,
which are shared across time steps and are used to selectively attend to the local and global evidence, respectively.
Given the local tokens $v^L_t$ and the global graph tokens $v^G_t$, we extract navigable evidence via cross-attention:
\begin{equation}
\tilde{\mathbf{Q}}^{\varphi }_{t}=\mathrm{CoAttn}(\mathbf{Q}^{\varphi }_{0},\, v^\varphi _t,\, v^\varphi _t),\quad \varphi \in \{L,G\},
\end{equation}
where $\mathrm{CoAttn}(\mathbf{Q},\mathbf{K},\mathbf{V})$ denotes multi-head attention taking $\mathbf{Q}$ as queries and $(\mathbf{K},\mathbf{V})$ as keys/values.
We further apply a lightweight refinement layer to obtain the final navigable queries:
\begin{equation}
\mathbf{Q}^{\varphi }_{t}=\tilde{\mathbf{Q}}^{\varphi }_{t}+\mathrm{FFN}(\tilde{\mathbf{Q}}^{\varphi }_{t}),\quad \varphi \in \{L,G\}.
\end{equation}
\textcolor{black}{
Here, we use the same learnable query slots $\mathbf{Q}^{\varphi}_{0}$ at every navigation step. 
At step $t$, the step-specific navigable queries $\mathbf{Q}^{\varphi}_{t}$ are recomputed by cross-attending $\mathbf{Q}^{\varphi}_{0}$ to the branch tokens $v^{\varphi}_{t}$. 
Because these branch tokens change with the current observation and the evolving graph, the generated queries change accordingly. 
Long-horizon context is maintained in the evolving global graph $G_t$ rather than in the initial query slots. 
Therefore, sharing $\mathbf{Q}^{\varphi}_{0}$ does not cause the loss of long-horizon intent. 
The resulting queries are further grounded with the full instruction at every step by the subsequent Instruction--Query Aligner.
}

\subsection{Instruction-Query Aligner (Align)}
We first concatenate the local and global navigable queries as $Q_t=[Q_t^L;Q_t^G]\in \mathbb{R}^{K_{\rm tot}\times d}$, where $K_{\rm tot}=K_L+K_G$.
To enhance the interaction between local and global evidence tokens, we apply a self-attention layer on $\mathbf{Q}_t$, followed by a feed-forward refinement with residual connection:
\begin{equation}
\hat{\mathbf{Q}}_t = \mathrm{MHSA}(\mathbf{Q}_t),\quad \hat{\mathbf{Q}}_t = \hat{\mathbf{Q}}_t + \mathrm{FFN}(\hat{\mathbf{Q}}_t).
\end{equation}
Let $\mathbf{P}\in\mathbb{R}^{M\times d}$ denote instruction representations processed by the language encoder.
We then perform cross-attention to fuse instruction semantics into the navigable evidence, taking navigable tokens as queries and instruction as keys/values:
\begin{equation}
\tilde{\mathbf{F}}_t = \hat{\mathbf{Q}}_t + \mathrm{CoAttn}(\hat{\mathbf{Q}}_t, \mathbf{P}, \mathbf{P}),
\end{equation}
\begin{equation}
\mathbf{F}^{\star}_t = \tilde{\mathbf{F}}_t + \mathrm{FFN}(\tilde{\mathbf{F}}_t).
\end{equation}
Finally, we apply an FFN to obtain the aligned cross-modal features $\mathbf{F}^{\star}_t$ which are used as the fused state for the subsequent policy head.

\subsection{Navigation-aware Token-adaptive Distillation (Distill)}
To obtain a compact student while preserving the teacher’s navigation intent, we perform distillation on the teacher’s action-sufficient local and global navigable queries at each time step. In addition, we distill the teacher’s action distribution via a policy KL loss during fine-tuning.

Let $\mathbf{Q}^{L}_{t,T}, \mathbf{Q}^{G}_{t,T}\in\mathbb{R}^{K_L\times d},\mathbb{R}^{K_G\times d}$ and
$\mathbf{Q}^{L}_{t,S}, \mathbf{Q}^{G}_{t,S}\in\mathbb{R}^{K_L\times d},\mathbb{R}^{K_G\times d}$ denote the teacher and student local/global navigable queries at step $t$.
We compute token-adaptive weights by measuring the teacher--student query alignment and emphasizing poorly aligned tokens. We first $\ell_2$-normalize queries along the feature dimension and compute the teacher--student similarity matrices:
\begin{equation}
\bar{\mathbf{Q}}^{L}_{t,\omega}=\mathrm{Norm}(\mathbf{Q}^{L}_{t,\omega}),\quad
\mathbf{S}^{L}_{t}=\bar{\mathbf{Q}}^{L}_{t,T}\big(\bar{\mathbf{Q}}^{L}_{t,S}\big)^{\top}\in\mathbb{R}^{K_L\times K_L},
\end{equation}
\begin{equation}
\bar{\mathbf{Q}}^{G}_{t,\omega}=\mathrm{Norm}(\mathbf{Q}^{G}_{t,\omega }),\quad
\mathbf{S}^{G}_{t}=\bar{\mathbf{Q}}^{G}_{t,T}\big(\bar{\mathbf{Q}}^{G}_{t,S}\big)^{\top}\in\mathbb{R}^{K_G\times K_G},
\end{equation}
where the dot ``$\omega$'' indicates either $T$ (teacher) or $S$ (student).
We then pool similarities to obtain a per-token best-match score and convert it into a \emph{difficulty} weight:
\begin{equation}
\mathbf{m}^{L}_{t}=\mathrm{Pool}\big(\mathbf{S}^{L}_{t}\big)\in\mathbb{R}^{K_L},\quad
\mathbf{m}^{G}_{t}=\mathrm{Pool}\big(\mathbf{S}^{G}_{t}\big)\in\mathbb{R}^{K_G},
\end{equation}
\begin{equation}
\mathbf{s}^{L}_{t}=1-\sigma(\mathbf{m}^{L}_{t}),\quad
\mathbf{s}^{G}_{t}=1-\sigma(\mathbf{m}^{G}_{t}),
\end{equation}
where $\mathrm{Pool}(\cdot)$ reduces each teacher token to a scalar via diagonal (index-wise) pooling, i.e., $\mathrm{Pool}(\mathbf{S})i=\mathbf{S}{ii}$; this is applicable because the teacher and student share the same query-slot design, making slot indices comparable across models. $\sigma(\cdot)$ is the sigmoid function.
Intuitively, if a teacher token is highly similar to its \emph{same-index} student token (large $\mathbf{m}$), its weight becomes small; otherwise, poorly aligned token pairs receive larger weights and are emphasized during distillation.
Finally, we normalize weights within each branch to stabilize training:
\begin{equation}
\tilde{\mathbf{s}}^{L}_{t}=\frac{\mathbf{s}^{L}_{t}}{\sum_{k=1}^{K_L} \mathbf{s}^{L}_{t,k}+\epsilon},\quad
\tilde{\mathbf{s}}^{G}_{t}=\frac{\mathbf{s}^{G}_{t}}{\sum_{k=1}^{K_G} \mathbf{s}^{G}_{t,k}+\epsilon}.
\end{equation}

With these token-adaptive weights, we distill the teacher queries to the student:
\begin{equation}
\begin{aligned}
\mathcal{L}_{\mathrm{NTD}}
= \sum_{t=1}^{T}\Bigg(
&\sum_{k=1}^{K_L}\tilde{\mathbf{s}}^{L}_{t,k}
\left\|\mathbf{q}^{L(k)}_{t,S}-\mathbf{q}^{L(k)}_{t,T}\right\|_2^2 \\
&+\sum_{k=1}^{K_G}\tilde{\mathbf{s}}^{G}_{t,k}
\left\|\mathbf{q}^{G(k)}_{t,S}-\mathbf{q}^{G(k)}_{t,T}\right\|_2^2
\Bigg),
\end{aligned}
\end{equation}
where $T$ denotes the number of decision steps in an episode (trajectory length).
During fine-tuning, we further distill the teacher's decision behavior by matching the action distributions of the teacher and the student:
\begin{equation}
\mathcal{L}_{\pi}=\sum_{t=1}^{T}\mathrm{KL}\Big(\pi_T(\cdot|t)\,\|\,\pi_S(\cdot|t)\Big),
\end{equation}
where $\mathrm{KL}(\cdot\|\cdot)$ denotes the Kullback--Leibler divergence.
$\pi_T(\cdot|t)$ and $\pi_S(\cdot|t)$ are the teacher and student action distributions over the discrete action set $\mathcal{A}_t=\mathcal{N}(v_t)\cup\{\texttt{STOP}\}$ at navigation step $t$.
This loss transfers the teacher's task-specific decision boundary and uncertainty to the student, complementing query-level distillation and improving fine-tuning stability.
During pretraining, we use only $\mathcal{L}_{\mathrm{NTD}}$ with unit weight. During fine-tuning, we balance $\mathcal{L}_{\mathrm{NTD}}$ and $\mathcal{L}_{\pi}$ by introducing a hyper-parameter $\alpha$, and optimize $\mathcal{L}_{D}
=\alpha\,\mathcal{L}_{\mathrm{NTD}}+(1-\alpha)\,\mathcal{L}_{\pi}$.

\section{Experiment}
% \subsection{Experimental Setup}
\subsection{Datasets and Evaluation Protocols}
We evaluate the proposed method on both fine-grained and goal-oriented datasets. The former (R2R~\cite{anderson2018vision} and RxR-English~\cite{ku2020room}) provide step-by-step navigation instructions, whereas the latter (REVERIE~\cite{qi2020reverie} and SOON~\cite{zhu2021soon}) additionally require identifying a target object at the destination. The R2R dataset consists of 21,576 instructions and 7,189 paths, divided into training, validation-seen, validation-unseen, and test-unseen sets. The RxR-English dataset extends the path in R2R, which is made up of 42,002 instructions. REVERIE provides concise instructions that describe target locations and objects. SOON has more complex instructions, including target areas, target objects, and neighboring areas, with an average instruction length of 47 words.

Following prior work, we evaluate the model's performance on the R2R dataset using success rate (SR), oracle SR (OSR), SR weighted by path length (SPL), and navigation error (NE). For the RxR dataset, we introduce SR weighted by dynamic time warping (sDTW) and normalized dynamic time warping (nDTW). For REVERIE and SOON, we add remote grounding success rate (RGS) and RGS weighted by path length (RGSPL).

\subsection{Implementation Details}
\label{app:implementation_details}
Following prior work~\cite{wang2024magic, wang2024causal,zhu2025minivln}, we use CLIP-B/16~\cite{pmlr-v139-radford21a} to extract panoramic image features and initialize the network weights from METER~\cite{Dou_2022_CVPR}.
Following the standard VLN pretraining recipe, we employ MLM~\cite{devlin-etal-2019-bert} and SAP~\cite{chen2021history} objectives for R2R and RxR, and additionally include OG~\cite{Lin_2021_CVPR} for REVERIE and SOON.
We adopt EnvEdit~\cite{Li_2022_CVPR} for feature augmentation and use the same synthetic extended datasets~\cite{Hao_2020_CVPR,10004992,Wang_2022_CVPR} as prior work for R2R/REVERIE/RxR.

\textbf{Teacher and student architectures.}
The teacher and student share the same overall topology with two branches of learnable query slots ($K_L=K_G=32$ for local/global queries), while the student is made compact by reducing the depth of the encoders and cross-modal reasoning modules.
For text and panoramic encoding, the teacher uses 6 and 2 transformer layers, respectively, whereas the student uses 1 and 1 layer.
In the Navigable Query Generator (NQG), we stack 4 cross-attention blocks for the teacher and 1 block for the student.
In the Instruction-Query Aligner (IQA), the teacher employs 3 layers of query self-attention and instruction--query cross-attention, while the student uses only 1 such layer.

\textbf{Optimization and schedules.}
Pretraining is conducted on a single NVIDIA L20 GPU using AdamW~\cite{2017Decoupled} for up to 300K iterations, with batch size 48 and learning rate $5\times10^{-5}$.
During fine-tuning, we use speaker models with environmental dropout to provide dynamic pseudo-labels.
We adopt batch size 12 for R2R and REVERIE, and batch size 5 for RxR and SOON, with learning rate $2\times10^{-5}$ and up to 100K iterations.
In both pretraining and fine-tuning, we follow a two-stage teacher--student procedure: we first train the teacher to convergence under the corresponding objectives, then freeze all teacher parameters and train the student with our distillation losses.
Unless otherwise stated, all remaining hyper-parameters follow the original training protocol. During the fine-tuning, we set the distillation hyper-parameter $\alpha=0.5$. 

\textbf{Parameter and GFLOPs Measurement Protocol.}
We computed the number of parameters (Params) and computational cost (GFLOPs) for all methods using their released codebases. GFLOPs were measured with the Python toolkit thop under a unified setting for fair comparison: a single-step forward inference with batch size = 8, instruction length = 44, 3 candidate nodes, and 6 previously visited (historical) nodes for every method. We report model forward-pass computation only, excluding any data pre-processing or post-processing overhead. For methods without publicly available implementations, we directly report the Params/GFLOPs numbers as stated in their original papers.

\textbf{On-device Throughput Measurement on Jetson Nano.}
We additionally evaluate practical on-device efficiency on a Jetson Nano.
Following the unified Params/GFLOPs protocol above, we deploy each model on Jetson Nano and measure \emph{Step FPS} (steps/s) under a single-step forward inference setting with batch size $=1$ (to reflect typical real-time embodied deployment).
We keep the input configuration consistent with the FLOPs setting (instruction length $=44$, 3 candidate nodes, and 6 previously visited nodes), and run inference in FP32.
The reported throughput measures the model forward-pass computation only, excluding any data pre-processing and post-processing overhead, and is computed as:
\begin{equation}
\mathrm{steps/s}=\frac{1}{T_{\mathrm{sec/step}}}=\frac{1000}{T_{\mathrm{ms/step}}},
\end{equation}
where $T_{\mathrm{ms/step}}$ denotes the average per-step forward latency (in milliseconds) measured on Jetson Nano.

% %%%%%%%%%%%%%%%%%%%%%%%%%%%%%%%%%%%%%%%%%%%%%%%%%%%%%%%%%%%%%%%%%%%%%%%%%%%%%%%
% %%%%%%%%%%%%%%%%%%%%%%%%%%%%%%%%%%%%%%%%%%%%%%%%%%%%%%%%%%%%%%%%%%%%%%%%%%%%%%%

\begin{table*}[!t]   
\centering
% \caption{Comparison with state-of-the-art methods on the R2R dataset. "–": unavailable statistics. Results highlighted in pink, yellow, and orange denote the best, second-best, and third-best methods, respectively. \textsuperscript{\textdaggerdbl} indicates methods whose Params and GFLOPs are measured by us using their publicly available code under the standard evaluation protocol.
% \textsuperscript{\textdagger} indicates methods without public code; for these, we directly report the Params/GFLOPs as stated in the original papers, which follow the same measurement protocol.}
\caption{\textcolor{black}{Comparison with state-of-the-art methods on R2R. 
Pink, yellow, and orange indicate the best, second-best, and third-best results, respectively, and ``--'' denotes unavailable results. 
$\ddagger$ marks Params/GFLOPs measured from released code under our protocol, whereas $\dagger$ marks values reported in the original papers under the same protocol.}}
\resizebox{\linewidth}{!}{
\begin{tabular}{l|cccc|cccc|cccc|cc}
\hline
\rowcolor[HTML]{F2F2F2} 
\multicolumn{1}{c|}{\cellcolor[HTML]{F2F2F2}}                                  & \multicolumn{4}{c|}{\cellcolor[HTML]{F2F2F2}\textbf{Validation   Seen}}                                                                                      & \multicolumn{4}{c|}{\cellcolor[HTML]{F2F2F2}\textbf{Validation   Unseen}}                                                                                 & \multicolumn{4}{c|}{\cellcolor[HTML]{F2F2F2}\textbf{Test   Unseen}}                                                                                          & \cellcolor[HTML]{F2F2F2}                                      & \cellcolor[HTML]{F2F2F2}                                   \\
\rowcolor[HTML]{F2F2F2} 
\multicolumn{1}{c|}{\multirow{-2}{*}{\cellcolor[HTML]{F2F2F2}\textbf{Method}}} & \textbf{SR$\uparrow$}                          & \textbf{SPL$\uparrow$}                         & \textbf{OSR$\uparrow$}                         & \textbf{NE$\downarrow$}                         & \textbf{SR$\uparrow$}                          & \textbf{SPL$\uparrow$}                         & \textbf{OSR$\uparrow$}                      & \textbf{NE$\downarrow$}                         & \textbf{SR$\uparrow$}                          & \textbf{SPL$\uparrow$}                         & \textbf{OSR$\uparrow$}                         & \textbf{NE$\downarrow$}                         & \multirow{-2}{*}{\cellcolor[HTML]{F2F2F2}\textbf{Param (M)↓}} & \multirow{-2}{*}{\cellcolor[HTML]{F2F2F2}\textbf{GFLOPs↓}} \\ \hline
HOP+\textsuperscript{\textdaggerdbl}\textsubscript{\cite{qiao2023hop_plus}}                                                                           & 78                                    & 73                                    & -                                     & 2.33                                 & 67                                    & 61                                    & -                                  & 3.49                                 & 66                                    & 60                                    & -                                     & 3.71                                 & 164.62                                                        & 50.77                                                      \\
KERM\textsuperscript{\textdaggerdbl}\textsubscript{\cite{li2023kerm}}                                                                           & 80                                    & 74                                    & -                                     & 2.19                                 & 72                                    & 61                                    & -                                  & 3.22                                 & 70                                    & 59                                    & -                                     & 3.61                                 & 222.04                                                        & 169.46                                                     \\
GeoVLN\textsuperscript{\textdaggerdbl}\textsubscript{\cite{huo2023geovln}}                                                                         & 79                                    & 76                                    & -                                     & 2.22                                 & 68                                    & 63                                    & -                                  & 3.35                                 & 65                                    & 61                                    & -                                     & 3.95                                 & 182.64                                                        & 296.65                                                     \\
DSRG\textsuperscript{\textdaggerdbl}\textsubscript{\cite{wang2023dual}}                                                                           & 81                                    & 76                                    & 88                                    & 2.23                                 & 73                                    & 62                                    & 81                                 & 3.00                                 & 72                                    & 61                                    & 78                                    & 3.33                                 & 188.87                                                        & 102.85                                                     \\
BEVBert\textsuperscript{\textdaggerdbl}\textsubscript{\cite{an2022bevbert}}                                                                        & 81                                    & 74                                    & 88                                    & 2.17                                 & 75                                    & 64                                    & 84                                 & 2.81                                 & 73                                    & 62                                    & 81                                    & 3.13                                 & 181.08                                                        & 294.32                                                     \\
GridMM\textsuperscript{\textdaggerdbl}\textsubscript{\cite{wang2023gridmm}}                                                                         & 80                                    & 74                                    & 85                                    & 2.34                                 & 75                                    & 64                                    & -                                  & 2.83                                 & 73                                    & 62                                    & -                                     & 3.35                                 & 161.00                                                        & 158.08                                                     \\
ScaleVLN\textsuperscript{\textdaggerdbl}\textsubscript{\cite{wang2023scaling}}                                                                       & 80                                    & 75                                    & 87                                    & 2.12                                 & 79                                    & 70                                    & \cellcolor[HTML]{FFCA9B} 87 & 2.34 & 77                                    & 68                                    & \cellcolor[HTML]{FFCA9B} 83                                    & \cellcolor[HTML]{FFCA9B} 2.73 & 181.02                                                        & 82.13                                                      \\
GOAT\textsuperscript{\textdaggerdbl}\textsubscript{\cite{wang2024causal}}                                                                           & 83.74 & \cellcolor[HTML]{FFCA9B} 79.48 & 88.64 & 1.79 & 77.82                                 & 68.13                                 & 84.72                              & 2.40                                 & 74.57                                 & 64.94                                 & 80.35                                 & 3.04                                 & 190.96                                                        & 65.98                                                      \\
MAGIC-L\textsuperscript{\textdagger}\textsubscript{\cite{wang2024magic}}                                                                        & \cellcolor[HTML]{FFCA9B} 83.84                                 & 79.33                                 & \cellcolor[HTML]{FFCA9B} 88.93                                 & \cellcolor[HTML]{FFCA9B} 1.73                                 & 78.88                                 & 69.82                                 & 86.12                              & 2.22                                 & 77.02                                 & \cellcolor[HTML]{FFCA9B} 68.73                                 & 81.88 & 2.75 & 190.96 & 65.98 \\
MAGIC-S\textsuperscript{\textdagger}\textsubscript{\cite{wang2024magic}}                                                                        & 78.16                                 & 71.10                                 & 85.41                                 & 2.37                                 & 76.03                                 & 65.07                                 & 84.89                              & 2.67                                 & 75.17                                 & 65.13                                 & 81.88 & 2.94                                 & \cellcolor[HTML]{FFCA9B} 11.14                         & \cellcolor[HTML]{FFCA9B} 2.03                       \\
3-D\textsubscript{\cite{tan2024self}}                                                                            & 80                                    & 73                                    & -                                     & 2.35                                 & 71                                    & 58                                    & -                                  & 3.36                                 & 70                                    & 58                                    & -                                     & 3.73                                 & -                                                             & -                                                          \\
RAM\textsubscript{\cite{wei2025unseen}}                                                                            & 82.47                                 & 77.98                                 & -                                     & 1.85                                 & 76.29                                 & 66.39                                 & -                                  & 2.66                                 & 75                                    & 65                                    & -                                     & 3.08                                 & -                                                             & -                                                          \\
EAM\textsubscript{\cite{tan2025source}}                                                                        & 78.86                                     & 73.39                                     & -                                     & 2.26                                    & 72.33 & 61.35 & -                                  & 3.23                                    & -  & - & -                                     & -                                    & -                                                         & -                                                          \\
Dual-SR\textsubscript{\cite{11246709}}                                                                        & 79                                     & 73                                     & -                                     & 2.30                                    & \cellcolor[HTML]{FFCA9B} 81 & \cellcolor[HTML]{FFCA9B} 76 & -                                  & \cellcolor[HTML]{FFCA9B} 2.20                                    & 70  & 61 & -                                     & 3.38                                    & -                                                         & -                                                          \\
DUET-Imagine\textsubscript{\cite{perincherry2025visual}}                                                                   & 79.90                                 & 73.75                                 & -                                     & 2.19                                 & 72.12                                 & 60.48                                 & -                                  & 3.19                                 & 71                                    & 60                                    & -                                     & 3.52                                 & -                                                             & -                                                          \\
MiniVLN\textsuperscript{\textdagger}\textsubscript{\cite{zhu2025minivln}}                                                                        & -                                     & -                                     & -                                     & -                                    &  78.80 & 70.17 & -                                  & -                                    & \cellcolor[HTML]{FFCA9B} 77.59 &  68.05 & -                                     & -                                    & 21.95                                                         & -                                                          \\ 
\hline
Ours-Teacher                                                                   & \cellcolor[HTML]{FFCCC9} 88.16                                 & \cellcolor[HTML]{FFCCC9} 84.41                                 & \cellcolor[HTML]{FFCCC9} 94.65                                 & \cellcolor[HTML]{FFCCC9} 1.43                                 & \cellcolor[HTML]{FFCCC9} 82.93                                 & \cellcolor[HTML]{FFCCC9} 78.12                                 & \cellcolor[HTML]{FFCCC9} 90.89                              & \cellcolor[HTML]{FFCCC9} 2.01                                 & \cellcolor[HTML]{FFCCC9} 80.43                                 & \cellcolor[HTML]{FFCCC9} 71.98                                 & \cellcolor[HTML]{FFCCC9} 86.16                                 & \cellcolor[HTML]{FFCCC9} 2.41                                 & \cellcolor[HTML]{FFCCC9} 150.78                                                        & \cellcolor[HTML]{FFCCC9} 45.56                                                      \\
Ours-Student                                                                   & \cellcolor[HTML]{FFFFD4} 85.97                                 & \cellcolor[HTML]{FFFFD4} 81.65                                 & \cellcolor[HTML]{FFFFD4} 90.12                                 & \cellcolor[HTML]{FFFFD4} 1.66                                 & \cellcolor[HTML]{FFFFD4} 81.19                                 & \cellcolor[HTML]{FFFFD4} 76.05                                 & \cellcolor[HTML]{FFFFD4} 88.74                              & \cellcolor[HTML]{FFFFD4} 2.11                                 & \cellcolor[HTML]{FFFFD4} 79.12                                 & \cellcolor[HTML]{FFFFD4} 69.54                                 & \cellcolor[HTML]{FFFFD4} 83.21                                 & \cellcolor[HTML]{FFFFD4} 2.57                                 & \cellcolor[HTML]{FFFFD4} 9.56                                                          & \cellcolor[HTML]{FFFFD4} 1.92                                                       \\ \hline
\end{tabular}
}

\label{tab:tab_r2r}

\end{table*}

\begin{table*}[!t]   
\centering
\caption{\textcolor{black}{Comparison with state-of-the-art methods on REVERIE. 
Pink, yellow, and orange indicate the best, second-best, and third-best results, respectively, and ``--'' denotes unavailable results.}}
\resizebox{\linewidth}{!}{
\begin{tabular}{l|cccc|cccc|cccc}
\hline
\rowcolor[HTML]{F2F2F2} 
\multicolumn{1}{c|}{\cellcolor[HTML]{F2F2F2}}                                  & \multicolumn{4}{c|}{\cellcolor[HTML]{F2F2F2}\textbf{Validation   Seen}} & \multicolumn{4}{c|}{\cellcolor[HTML]{F2F2F2}\textbf{Validation   Unseen}} & \multicolumn{4}{c}{\cellcolor[HTML]{F2F2F2}\textbf{Test   Unseen}} \\
\rowcolor[HTML]{F2F2F2} 
\multicolumn{1}{c|}{\multirow{-2}{*}{\cellcolor[HTML]{F2F2F2}\textbf{Method}}} & \textbf{SR↑}    & \textbf{SPL↑}   & \textbf{RGS↑}   & \textbf{RGSPL↑}   & \textbf{SR↑}    & \textbf{SPL↑}    & \textbf{RGS↑}    & \textbf{RGSPL↑}   & \textbf{SR↑}  & \textbf{SPL↑}  & \textbf{RGS↑}  & \textbf{RGSPL↑}  \\ \hline
BEVBert\textsubscript{\cite{an2022bevbert}}                                      & 73.72   & 65.32   & 57.70   & 51.73    & 51.78    & 36.37    & 34.71   & 24.44    & 52.81  & 36.41  & 32.06  & 22.09  \\
GOAT\textsubscript{\cite{wang2024causal}}                                         & 78.64   & \cellcolor[HTML]{FFCA9B} 71.40   & \cellcolor[HTML]{FFCA9B} 63.74   & \cellcolor[HTML]{FFCA9B} 57.85    & 53.37    & 36.70    & \cellcolor[HTML]{FFCA9B} 38.43   & 26.09    & \cellcolor[HTML]{FFCA9B} 57.72  & \cellcolor[HTML]{FFCA9B} 40.53  & \cellcolor[HTML]{FFCA9B} 38.32  & \cellcolor[HTML]{FFCA9B} 26.70  \\
EAM\textsubscript{\cite{tan2025source}}                                     & 72.59   & 64.96   & 57.83   & 51.80    & 48.00    & 35.65    & 32.72   & 24.22    & -      & -      & -      & -      \\
WPCL\textsubscript{\cite{wang2025weakly}}                                         & \cellcolor[HTML]{FFCA9B} 79.71   & 71.39   & -       & -        & 54.03    & 37.41    & -       & -        & -      & -      & -      & -      \\
CONSOLE\textsubscript{\cite{lin2024correctable}}                                      & 74.14   & 65.15   & 60.08   & 52.69    & 50.07    & 34.40    & 34.05   & 23.33    & 55.13  & 37.13  & 33.18  & 22.25  \\
NavQ\textsubscript{\cite{xu2025navq}}                                         & -       & -       & -       & -        & 54.10    & 39.22    & 37.57   & \cellcolor[HTML]{FFCA9B} 27.29    & 54.91  & 40.08  & 35.87  & 25.14  \\
MiniVLN\textsubscript{\cite{zhu2025minivln}}                                      & -       & -       & -       & -        & 54.30    & \cellcolor[HTML]{FFCA9B} 42.02    & 35.16   & 27.06    & 53.62  & 39.66  & 31.18  & 23.33  \\
VLN-VER\textsubscript{\cite{liu2024volumetric}}                                      & 75.83   & 66.19   & 61.71   & 56.20    & \cellcolor[HTML]{FFCA9B} 55.98    & 39.66    & 33.71   & 23.70    & 56.82  & 38.76  & 33.88  & 23.19  \\ \hline
Ours-Teacher                                 & \cellcolor[HTML]{FFCCC9} 83.21   & \cellcolor[HTML]{FFCCC9} 74.66   & \cellcolor[HTML]{FFCCC9} 67.98   & \cellcolor[HTML]{FFCCC9} 61.23    & \cellcolor[HTML]{FFCCC9} 58.97    & \cellcolor[HTML]{FFCCC9} 46.61    & \cellcolor[HTML]{FFCCC9} 42.42   & \cellcolor[HTML]{FFCCC9} 31.02    & \cellcolor[HTML]{FFCCC9} 60.19  & \cellcolor[HTML]{FFCCC9} 44.21  & \cellcolor[HTML]{FFCCC9} 41.92  & \cellcolor[HTML]{FFCCC9} 29.61  \\
Ours-Student                                 & \cellcolor[HTML]{FFFFD4} 81.03   & \cellcolor[HTML]{FFFFD4} 72.44   & \cellcolor[HTML]{FFFFD4} 66.01   & \cellcolor[HTML]{FFFFD4} 59.43    & \cellcolor[HTML]{FFFFD4} 57.01    & \cellcolor[HTML]{FFFFD4} 45.92    & \cellcolor[HTML]{FFFFD4} 41.01   & \cellcolor[HTML]{FFFFD4} 29.98    & \cellcolor[HTML]{FFFFD4} 59.03  & \cellcolor[HTML]{FFFFD4} 43.12  & \cellcolor[HTML]{FFFFD4} 40.31  & \cellcolor[HTML]{FFFFD4} 28.23  \\ \hline
\end{tabular}
}
\label{tab:tab_REVERIE}
\end{table*}

\begin{table}[!t]   
\centering
% \caption{Comparison with state-of-the-art methods on the SOON dataset. "–": unavailable statistics. Results highlighted in pink, yellow, and orange denote the best, second-best, and third-best methods, respectively.}
\caption{\textcolor{black}{Comparison with state-of-the-art methods on SOON. 
Pink, yellow, and orange indicate the best, second-best, and third-best results, respectively, and ``--'' denotes unavailable results.}}
\resizebox{\linewidth}{!}{
\begin{tabular}{l|cccc|cccc}
\hline
\rowcolor[HTML]{F2F2F2} 
\multicolumn{1}{c|}{\cellcolor[HTML]{F2F2F2}}                                  & \multicolumn{4}{c|}{\cellcolor[HTML]{F2F2F2}\textbf{Validation   Unseen}}                                                                                    & \multicolumn{4}{c}{\cellcolor[HTML]{F2F2F2}\textbf{Test   Unseen}}                                                                                           \\
\rowcolor[HTML]{F2F2F2} 
\multicolumn{1}{c|}{\multirow{-2}{*}{\cellcolor[HTML]{F2F2F2}\textbf{Method}}} & \textbf{SR↑}                          & \textbf{SPL↑}                         & \textbf{OSR↑}                         & \textbf{RGSPL↑}                      & \textbf{SR↑}                          & \textbf{SPL↑}                         & \textbf{OSR↑}                         & \textbf{RGSPL↑}                      \\ \hline
DUET\textsubscript{\cite{chen2022think}}                                                                           & 36.28                                 & 22.58                                 & 50.91                                 & 3.75                                 & 33.44                                 & 21.42                                 & 43.00                                 & 4.17                                 \\
GridMM\textsubscript{\cite{wang2023gridmm}}                                                                         & 37.46                                 & 24.81                                 & 53.39                                 & 3.91                                 & 36.27                                 & 21.25                                 & 48.02                                 & 4.15                                 \\
GOAT\textsubscript{\cite{wang2024causal}}                                                                           & 40.35                                 & \cellcolor[HTML]{FFCA9B} 28.05 & 54.69                                 & \cellcolor[HTML]{FFCA9B} 5.85 & \cellcolor[HTML]{FFCA9B} 40.50 & \cellcolor[HTML]{FFCA9B} 25.18 & \cellcolor[HTML]{FFCA9B} 50.63 & \cellcolor[HTML]{FFCA9B} 6.10 \\
WPCL\textsubscript{\cite{wang2025weakly}}                                                                           & \cellcolor[HTML]{FFCA9B} 42.92 & 26.57                                 & 52.21                                 & -                                    & -                                     & -                                     & -                                     & -                                    \\
NavQ\textsubscript{\cite{xu2025navq}}                                                                           & 39.09                                 & 26.65                                 & \cellcolor[HTML]{FFCA9B} 58.79 & 5.51                                 & 38.59                                 & 24.50                                 & 48.92                                 & 4.48                                 \\ \hline
Ours-Teacher                                                                   & \cellcolor[HTML]{FFCCC9} 45.67                                 & \cellcolor[HTML]{FFCCC9} 31.59                                 & \cellcolor[HTML]{FFCCC9} 62.33                                 & \cellcolor[HTML]{FFCCC9} 6.94                                 & \cellcolor[HTML]{FFCCC9} 43.42                                 & \cellcolor[HTML]{FFCCC9} 28.03                                 & \cellcolor[HTML]{FFCCC9} 53.44                                 & \cellcolor[HTML]{FFCCC9} 7.31                                 \\
Ours-Student                                                                   & \cellcolor[HTML]{FFFFD4} 43.89                                 & \cellcolor[HTML]{FFFFD4} 30.02                                 & \cellcolor[HTML]{FFFFD4} 61.02                                 & \cellcolor[HTML]{FFFFD4} 6.01                                 & \cellcolor[HTML]{FFFFD4} 41.03                                 & \cellcolor[HTML]{FFFFD4} 27.11                                 & \cellcolor[HTML]{FFFFD4} 51.34                                 & \cellcolor[HTML]{FFFFD4} 6.97                                 \\ \hline
\end{tabular}
}
\label{tab:tab_soon}
\end{table}

\subsection{Comparison with State-of-the-Art methods}
\subsubsection{Comparison of effectiveness}
\textcolor{black}{Tab.}~\ref{tab:tab_r2r}--\ref{tab:tab_rxr} compare our method with prior state-of-the-art approaches on R2R, REVERIE, RxR-English, and SOON, respectively. 
Overall, our teacher model consistently delivers strong navigation performance and accurate object grounding across both seen and unseen environments. More importantly, benefiting from the proposed Navigation-aware Token-adaptive Distillation and policy distillation, our compact student effectively inherits the teacher’s capabilities and achieves competitive results, ranking among the top-performing methods on all four benchmarks.

For example, on R2R, our student improves SR over MiNiVLN~\cite{zhu2025minivln} with relative gains of 2.39\% and 1.53\% on the two evaluation splits. On REVERIE, the student achieves substantial improvements in RGSPL, with relative gains of 2.92\% and 4.90\% on the two splits. On the more challenging RxR and SOON benchmarks, our student also yields consistent improvements across key metrics, demonstrating strong robustness and generalization compared with previous methods.

\begin{table}[!t]   
\centering
% \caption{Comparison with state-of-the-art methods on the RxR-English dataset. "–": unavailable statistics. Results highlighted in pink, yellow, and orange denote the best, second-best, and third-best methods, respectively.}
\caption{\textcolor{black}{Comparison with state-of-the-art methods on RxR-English. 
Pink, yellow, and orange indicate the best, second-best, and third-best results, respectively, and ``--'' denotes unavailable results.}}
\resizebox{\linewidth}{!}{
\begin{tabular}{l|cccc|cccc}
\hline
\rowcolor[HTML]{F2F2F2} 
\multicolumn{1}{c|}{\cellcolor[HTML]{F2F2F2}}                                  & \multicolumn{4}{c|}{\cellcolor[HTML]{F2F2F2}\textbf{Validation   Seen}}                                                                                   & \multicolumn{4}{c}{\cellcolor[HTML]{F2F2F2}\textbf{Validation   Unseen}}                                                                                  \\
\rowcolor[HTML]{F2F2F2} 
\multicolumn{1}{c|}{\multirow{-2}{*}{\cellcolor[HTML]{F2F2F2}\textbf{Method}}} & \textbf{SR$\uparrow$}                         & \textbf{SPL$\uparrow$}                        & \textbf{nDTW$\uparrow$}                       & \textbf{sDTW$\uparrow$}                       & \textbf{SR$\uparrow$}                         & \textbf{SPL$\uparrow$}                        & \textbf{nDTW$\uparrow$}                       & \textbf{sDTW$\uparrow$}                       \\ \hline
HOP+\textsubscript{\cite{qiao2023hop_plus}}                                                                           & 53.6                                 & 47.9                                 & 59.0                                 & 43.0                                 & 45.7                                 & 38.4                                 & 52.0                                 & 36.0                                 \\
ADAPT\textsubscript{\cite{lin2022adapt}}                                                                          & 50.3                                 & 44.6                                 & 56.3                                 & 40.6                                 & 46.9                                 & 40.2                                 & 54.1                                 & 37.7                                 \\
CLEAR-C\textsubscript{\cite{li2022clear}}                                                                        & -                                    & -                                    & -                                    & -                                    & 46.0                                 & 40.1                                 & 57.2                                 & 38.7                                 \\
MAR\textsubscript{M-MP}\textsubscript{\cite{kamath2023new}}                                                                        & -                                    & -                                    & -                                    & -                                    & 50.2                                 & -                                    & 60.3                                 & 43.9                                 \\
PETL\textsubscript{\cite{qiao2023vln}}                                                                           & 60.5                                 & 56.8                                 & 65.7                                 & 51.7                                 & 57.9                                 & 54.2                                 & 64.9                                 & 49.7                                 \\
GOAT\textsubscript{\cite{wang2024causal}}                                                                           & \cellcolor[HTML]{FFCA9B} 82.0 & 76.7                                 & 76.2                                 & 68.9                        & 70.8                                 & 61.8                                 & 66.8                                 & 56.7                                 \\
MAGIC-L\textsubscript{\cite{wang2024magic}}                                                                        & 81.3                                 & \cellcolor[HTML]{FFCA9B} 77.5 & \cellcolor[HTML]{FFCA9B} 76.6 & \cellcolor[HTML]{FFCA9B} 69.2 & \cellcolor[HTML]{FFCA9B} 72.9 & \cellcolor[HTML]{FFCA9B} 65.4 & \cellcolor[HTML]{FFCA9B} 68.1 & \cellcolor[HTML]{FFCA9B} 58.7 \\
MAGIC-S\textsubscript{\cite{wang2024magic}}                                                                        & 70.2                                 & 63.8                                 & 65.7                                 & 56.0                                 & 68.1                                 & 59.8                                 & 64.0                                 & 53.6                                 \\
Dual-SR\textsubscript{\cite{11246709}}                                                                        & 62.1                                 & -                                    & 65.1                                 & -                                    & 64.9                                 & -                                    & \cellcolor[HTML]{FFCA9B} 68.1 & -                                    \\
EAM\textsubscript{\cite{tan2025source}}                                                                       & 60.1                                 & 56.5                                 & 66.0                                 & 51.5                                 & 57.5                                 & 53.7                                 & 64.2                                 & 49.1                                 \\ \hline
Ours-Teacher                                                                   & \cellcolor[HTML]{FFCCC9} 86.1                                 & \cellcolor[HTML]{FFCCC9} 81.9                                 & \cellcolor[HTML]{FFCCC9} 81.1                                 & \cellcolor[HTML]{FFCCC9} 71.1                                 & \cellcolor[HTML]{FFCCC9} 77.1                                 & \cellcolor[HTML]{FFCCC9} 69.9                                 & \cellcolor[HTML]{FFCCC9} 73.6                                 & \cellcolor[HTML]{FFCCC9} 63.1                                 \\
Ours-Student                                                                   & \cellcolor[HTML]{FFFFD4} 83.4                                 & \cellcolor[HTML]{FFFFD4} 79.1                                 & \cellcolor[HTML]{FFFFD4} 78.2                                 & \cellcolor[HTML]{FFFFD4} 70.9                                 & \cellcolor[HTML]{FFFFD4} 73.4                                 & \cellcolor[HTML]{FFFFD4} 67.2                                 & \cellcolor[HTML]{FFFFD4} 69.9                                 & \cellcolor[HTML]{FFFFD4} 61.2                                 \\ \hline
\end{tabular}
}
\label{tab:tab_rxr}
\end{table}

\subsubsection{Multi-seed stability analysis.}
We additionally evaluate the stability of QAD by reporting the mean and standard deviation over five runs with different random seeds. 
As reported in \textcolor{black}{Tab.}~\ref{tab:teacher_student_r2r_rxr} and \textcolor{black}{Tab.}~\ref{tab:teacher_student_reverie_soon}, the standard deviations are consistently small across R2R, RxR-English, REVERIE, and SOON, showing that both the teacher and student are insensitive to random initialization. 
More importantly, the compact student consistently approaches the teacher across all benchmarks. 
For example, on the R2R test-unseen, the student achieves 79.10 SR and 69.53 SPL, only slightly lower than the teacher's 80.43 SR and 71.98 SPL. 
On REVERIE and SOON, the student also retains most of the teacher's navigation and grounding ability, confirming that the proposed distillation strategy can stably transfer navigation-aware evidence selection from the teacher to the compact student.

\begin{table*}[!t]   
\centering
% \caption{\textbf{Teacher vs. student performance with multiple seeds.}
% Results on R2R (\textit{test-unseen}) and RxR-English (\textit{val-unseen}).
% All numbers are reported as mean$\pm$std over 5 runs with seeds \{0,1,2,3,4\}.}
\caption{\textcolor{black}{Teacher--student performance on R2R test-unseen and RxR-English validation-unseen. Values are reported as mean $\pm$ standard deviation over five runs with seeds $\{0,1,2,3,4\}$.}}
\resizebox{\linewidth}{!}{
\begin{tabular}{c|cccc|cccc}
\hline
\rowcolor[HTML]{F2F2F2} 
\cellcolor[HTML]{F2F2F2}                                  & \multicolumn{4}{c|}{\cellcolor[HTML]{F2F2F2}\textbf{Test   Unseen (R2R)}} & \multicolumn{4}{c}{\cellcolor[HTML]{F2F2F2}\textbf{Validation Unseen   (RxR-English)}} \\
\rowcolor[HTML]{F2F2F2} 
\multirow{-2}{*}{\cellcolor[HTML]{F2F2F2}\textbf{Method}} & \textbf{SR↑}     & \textbf{SPL↑}     & \textbf{OSR↑}    & \textbf{NE↓}    & \textbf{SR↑}       & \textbf{SPL↑}       & \textbf{nDTW↑}       & \textbf{sDTW↑}       \\ \hline
Ours-Teacher                                              & 80.43$\pm$0.04     & 71.98$\pm$0.02      & 86.14$\pm$0.04     & 2.40$\pm$0.03     & 77.1$\pm$0.01        & 69.8$\pm$0.2          & 73.4$\pm$0.3           & 63.0$\pm$0.2           \\
Ours-Student                                              & 79.10$\pm$0.05     & 69.53$\pm$0.03      & 83.21$\pm$0.05     & 2.57$\pm$0.01     & 73.1$\pm$0.02        & 67.2$\pm$0.1          & 69.8$\pm$0.2           & 61.2$\pm$0.1           \\ \hline
\end{tabular}
}
\label{tab:teacher_student_r2r_rxr}
\end{table*}

\begin{table*}[!t]   
\centering
% \caption{\textbf{Teacher vs. student performance on goal-oriented VLN.}
% Results on REVERIE and SOON (\textit{test-unseen}).
% All numbers are reported as mean$\pm$std over 5 runs with seeds \{0,1,2,3,4\}.}
\caption{\textcolor{black}{Teacher--student performance on the goal-oriented REVERIE and SOON test-unseen splits. Values are reported as mean $\pm$ standard deviation over five runs with seeds $\{0,1,2,3,4\}$.}}
\resizebox{\linewidth}{!}{
\begin{tabular}{c|cccc|cccc}
\hline
\rowcolor[HTML]{F2F2F2} 
\cellcolor[HTML]{F2F2F2}                                  & \multicolumn{4}{c|}{\cellcolor[HTML]{F2F2F2}\textbf{Test   Unseen (REVERIE)}} & \multicolumn{4}{c}{\cellcolor[HTML]{F2F2F2}\textbf{Test   Unseen (SOON)}} \\
\rowcolor[HTML]{F2F2F2} 
\multirow{-2}{*}{\cellcolor[HTML]{F2F2F2}\textbf{Method}} & \textbf{SR↑}     & \textbf{SPL↑}     & \textbf{RGS↑}     & \textbf{RGSPL↑}    & \textbf{SR↑}    & \textbf{SPL↑}    & \textbf{OSR↑}    & \textbf{RGSPL↑}   \\ \hline
Ours-Teacher                                              & 60.18$\pm$0.04     & 44.20$\pm$0.02      & 41.90$\pm$0.02      & 29.62$\pm$0.02       & 43.43$\pm$0.01    & 28.04$\pm$0.01     & 53.45$\pm$0.04     & 7.32$\pm$0.02       \\
Ours-Student                                              & 59.01$\pm$0.03     & 43.12$\pm$0.01      & 40.32$\pm$0.01      & 28.24$\pm$0.03       & 41.04$\pm$0.01    & 27.13$\pm$0.02    & 51.36$\pm$0.02     & 6.98$\pm$0.03       \\ \hline
\end{tabular}
}
\label{tab:teacher_student_reverie_soon}
\end{table*}

\begin{table*}[!t]   
\centering
% \caption{Controlled comparison of teacher-side cross-modal
% interaction designs on R2R and SOON test-unseen. All variants
% use the same local/global input representations, encoder
% configuration, hidden dimension, policy head, training data,
% and optimization schedule; only the interaction interface is
% changed. }
\caption{\textcolor{black}{Controlled comparison of teacher-side cross-modal interaction designs on R2R and SOON test-unseen. 
All non-interaction components and training settings are held fixed.}}
\label{tab:interaction_design}
\resizebox{\linewidth}{!}{
\begin{tabular}{l|cccc|cccc}
\hline

\rowcolor[HTML]{F2F2F2}
\cellcolor[HTML]{F2F2F2}
& \multicolumn{4}{c|}{\cellcolor[HTML]{F2F2F2}\textbf{R2R (Test Unseen)}}
& \multicolumn{4}{c}{\cellcolor[HTML]{F2F2F2}\textbf{SOON (Test Unseen)}} \\

\rowcolor[HTML]{F2F2F2}
\multirow{-2}{*}{\cellcolor[HTML]{F2F2F2}\textbf{Interaction design}}
& \textbf{SR$\uparrow$}
& \textbf{SPL$\uparrow$}
& \textbf{OSR$\uparrow$}
& \textbf{NE$\downarrow$}
& \textbf{SR$\uparrow$}
& \textbf{SPL$\uparrow$}
& \textbf{OSR$\uparrow$}
& \textbf{RGSPL$\uparrow$} \\ \hline

\textcolor{black}{DUET-style dense interaction}
& \textcolor{black}{77.91} & \textcolor{black}{68.12} & \textcolor{black}{84.31} & \textcolor{black}{2.68}
& \textcolor{black}{40.02} & \textcolor{black}{25.47} & \textcolor{black}{50.03} & \textcolor{black}{6.73} \\

\textcolor{black}{Shared generic queries}
& \textcolor{black}{79.21} & \textcolor{black}{69.21} & \textcolor{black}{84.99} & \textcolor{black}{2.54}
& \textcolor{black}{40.54} & \textcolor{black}{25.68} & \textcolor{black}{50.74} & \textcolor{black}{6.95} \\

\rowcolor[HTML]{effafc}
\textcolor{black}{QAD (Full)}
& \textcolor{black}{80.43} & \textcolor{black}{71.98} & \textcolor{black}{86.16} & \textcolor{black}{2.41}
& \textcolor{black}{43.42} & \textcolor{black}{28.03} & \textcolor{black}{53.44} & \textcolor{black}{7.31} \\ \hline

\end{tabular}
}
\end{table*}

\begin{figure}[t]
\centering
\includegraphics[width=\linewidth]{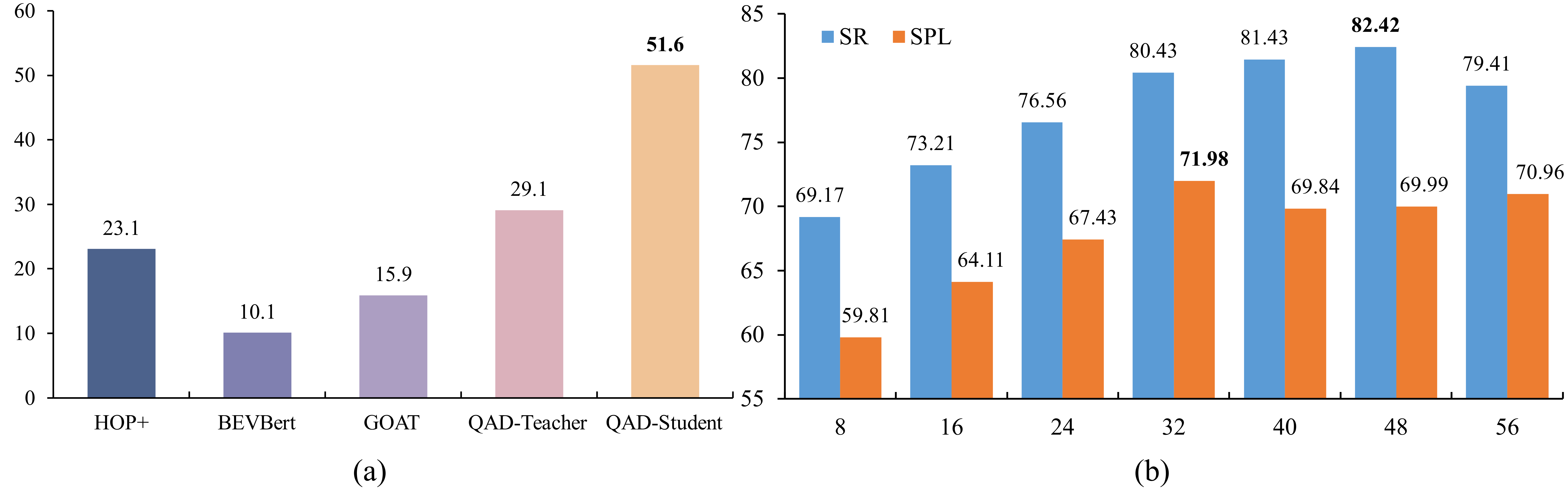}
\caption{\textcolor{black}{Jetson runtime and query-number ablations on R2R.
(a) Step throughput (steps/s) measured on Jetson Nano for representative VLN agents, including HOP+, BEVBert, GOAT, and our QAD teacher/student.
(b) R2R test-unseen SR and SPL versus the number of navigable queries per branch $K_b$, where $K_L=K_G=K_b$}}
\label{fig:ablation1}
\end{figure}

\subsubsection{Comparison of Efficiency}
To assess model size and computational cost, we report the number of parameters (Param) and computational cost (GFLOPs) following the same evaluation protocol and configuration as those in~\cite{wang2024magic,10801563}. 
As shown in \textcolor{black}{Tab.}~\ref{tab:tab_r2r}, our student model achieves the lowest parameter count and GFLOPs among all compared methods, indicating a markedly improved accuracy--efficiency trade-off. 
This improvement stems from (i) an action-sufficient query bottleneck that enables cross-modal reasoning on only K tokens instead of dense panoramic or graph features, and (ii) a substantially shallower student (with fewer text/panorama encoder layers and NQG/IQA layers) trained via Navigation-aware Token-adaptive Distillation and policy distillation, allowing it to retain the teacher’s capabilities while significantly reducing computation and memory usage.

We further report real-device runtime on an edge platform.
Specifically, \textcolor{black}{Fig.}~\ref{fig:ablation1}(a) shows the step throughput (steps/s) measured on Jetson Nano for representative VLN methods, including HOP+, BEVBert, GOAT, and our QAD teacher/student, under the same evaluation pipeline.
Despite being a stronger model, our QAD-Teacher already runs faster than GOAT (29.1 vs.\ 15.9 steps/s), reflecting the benefit of the action-sufficient query bottleneck that avoids dense panoramic/graph-level cross-modal computation.
More importantly, our compact student achieves the highest throughput (51.6 steps/s), yielding about $1.8\times$ speedup over the teacher and $3.2\times$ over GOAT.

\begin{figure}[t]
\centering
\includegraphics[width=\linewidth]{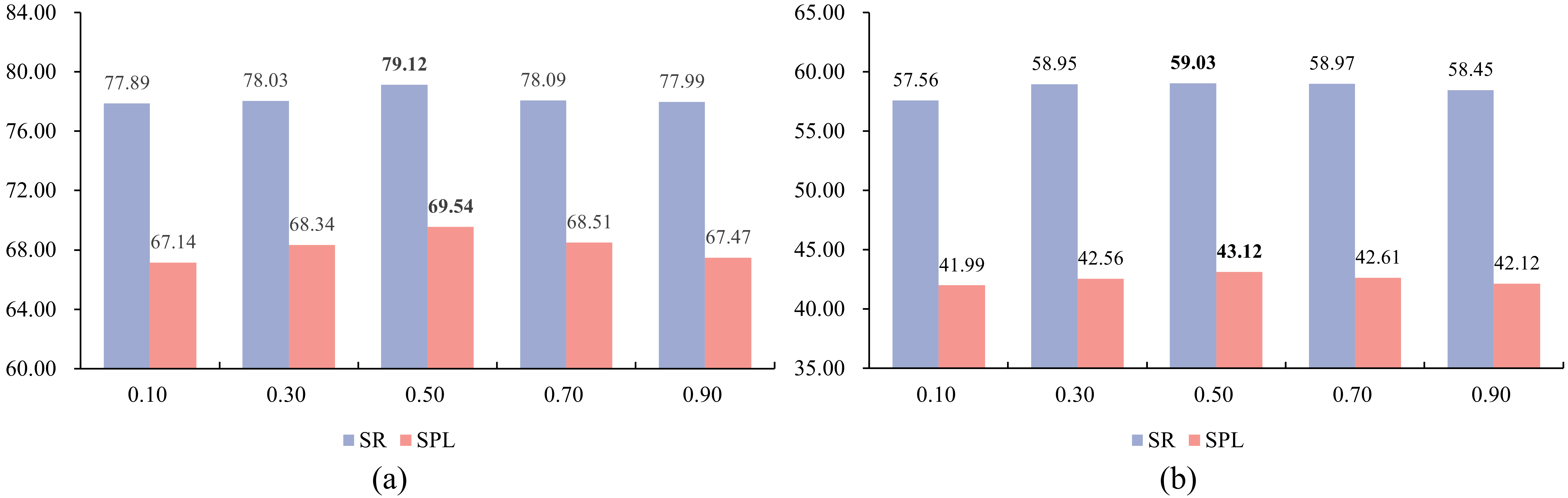}
\caption{\textcolor{black}{Sensitivity to the distillation hyper-parameter $\alpha$.
We sweep the $\alpha$ that balances query-level distillation $\mathcal{L}_{\mathrm{NTD}}$ and policy KL distillation $\mathcal{L}_{\pi}$ during \emph{fine-tuning}:
$\mathcal{L}_{D}=\alpha\,\mathcal{L}_{\mathrm{NTD}}+(1-\alpha)\,\mathcal{L}_{\pi}$.
Results are reported on R2R (\textit{test-unseen}: SR/SPL) and REVERIE (\textit{test-unseen}: SR/SPL).
A moderate $\alpha$ yields the best trade-off, while overly small/large $\alpha$ over-emphasizes policy- or query-level supervision, respectively.}}
\label{fig:alpha_ablation}
\end{figure}

\subsection{Ablation Studies}
We conduct ablation studies on the R2R and SOON test-unseen splits to analyze the key design choices of our
teacher--student framework, including (i) controlled comparisons of cross-modal interaction interfaces, (ii) core modules and distillation objectives, (iii) architecture scaling along encoder and cross-modal reasoning depth, (iv) distillation targets across local/global query branches, and (v) efficiency and the effect of the query-slot number.

\subsubsection{Controlled comparison of cross-modal interaction designs.}
\textcolor{black}{
Table~\ref{tab:interaction_design} compares different interaction interfaces while retaining the same local/global input representations, encoder configuration, policy head, training data, and optimization protocol. 
The DUET-style baseline directly performs instruction grounding over the original local/global branch tokens. 
The shared-query baseline introduces the same total number of learnable queries as QAD, but uses a single branch-agnostic query bank over the concatenated local/global evidence. 
QAD instead uses separate local and global query banks with navigation-specific semantic roles.
}

\textcolor{black}{
Compared with DUET-style dense interaction, QAD improves SR and SPL by 2.52 and 3.86 points on R2R, respectively, and by 3.40 and 2.56 points on SOON. 
Introducing shared generic queries produces only modest gains over dense interaction: 1.30/1.09 SR/SPL points on R2R and 0.52/0.21 points on SOON. 
With the same total number of 64 queries, the full QAD further improves SR/SPL over the shared-query baseline by 1.22/2.77 points on R2R and 2.88/2.35 points on SOON.
}

\textcolor{black}{
These results show that the improvement does not arise merely from inheriting a dual-branch backbone or inserting generic learnable query slots. 
Explicitly separating immediate local actionability from long-horizon global topological evidence provides a more effective cross-modal reasoning interface.
}

\begin{table}[!t]   
\centering
% \caption{\textbf{Ablation of core components and distillation objectives on R2R (Test Unseen).}
% We analyze NQG, IQA, NTD, policy KL, token-adaptive weighting, and compare NTD with a controlled attention-, feature-, and logit-level distillation baseline following the target types used in MAGIC~\cite{wang2024magic}.}
\caption{\textcolor{black}{Component and distillation ablations on R2R test-unseen.}}
\resizebox{\linewidth}{!}{
\begin{tabular}{c|l|cccc}
\hline
\rowcolor[HTML]{F2F2F2} 
\cellcolor[HTML]{F2F2F2}                                  & \multicolumn{1}{c|}{\cellcolor[HTML]{F2F2F2}}                                 & \multicolumn{4}{c}{\cellcolor[HTML]{F2F2F2}\textbf{Test   Unseen}} \\
\rowcolor[HTML]{F2F2F2} 
\multirow{-2}{*}{\cellcolor[HTML]{F2F2F2}\textbf{Option}} & \multicolumn{1}{c|}{\multirow{-2}{*}{\cellcolor[HTML]{F2F2F2}\textbf{Model}}} & \textbf{SR↑}   & \textbf{SPL↑}   & \textbf{OSR↑}   & \textbf{NE↓}  \\ \hline
\rowcolor[HTML]{effafc}
1                                                         & Teacher (full)                                                                & 80.43          & 71.98           & 86.16           & 2.41          \\
2                                                         & Teacher w/o NQG                                                               & 78.69          & 68.67           & 84.91           & 2.61          \\
3                                                         & Teacher w/o IQA                                                               & 77.13          & 67.92           & 83.11           & 2.74          \\
\rowcolor[HTML]{effafc}
4                                                         & Student (full)                                                                & 79.12          & 69.54           & 83.21           & 2.57          \\
5                                                         & Student w/o NTD                                                               & 70.23          & 61.44           & 72.14           & 3.71          \\
6                                                         & Student w/o Policy KL                                                         & 76.17          & 66.13           & 80.88           & 2.94          \\
7                                                         & Student w/o distillation                                                         & 67.21          & 61.40           & 72.39           & 3.75          \\
8                                                         & Student w/ uniform weights                                                    & 77.52          & 66.01           & 81.14           & 2.76          \\ 
\textcolor{black}{9}                                                         & \textcolor{black}{Attn.+Feat.+Logit KD~\cite{wang2024magic}}                                                    & \textcolor{black}{78.31}          & \textcolor{black}{66.54}           & \textcolor{black}{81.52}           & \textcolor{black}{2.69}          \\
\hline
\end{tabular}
}
\label{tab:ablate_core1}
\end{table}

\begin{table}[!t]   
\centering
% \caption{\textbf{Ablation of core components and distillation objectives on SOON (Test Unseen).}
% We analyze NQG, IQA, NTD, policy KL, token-adaptive weighting, and compare NTD with a controlled attention-, feature-, and logit-level distillation baseline following the target types used in MAGIC~\cite{wang2024magic}.}
\caption{\textcolor{black}{Component and distillation ablations on SOON test-unseen.}}
\resizebox{\linewidth}{!}{
\begin{tabular}{c|l|cccc}
\hline
\rowcolor[HTML]{F2F2F2} 
\cellcolor[HTML]{F2F2F2}                                  & \multicolumn{1}{c|}{\cellcolor[HTML]{F2F2F2}}                                 & \multicolumn{4}{c}{\cellcolor[HTML]{F2F2F2}\textbf{Test   Unseen}} \\
\rowcolor[HTML]{F2F2F2} 
\multirow{-2}{*}{\cellcolor[HTML]{F2F2F2}\textbf{Option}} & \multicolumn{1}{c|}{\multirow{-2}{*}{\cellcolor[HTML]{F2F2F2}\textbf{Model}}} & \textbf{SR↑}   & \textbf{SPL↑}   & \textbf{OSR↑}   & \textbf{RGSPL↑}  \\ \hline
\rowcolor[HTML]{effafc}
1                                                         & Teacher (full)                                                                & 43.42          & 28.03           & 53.44           & 7.31          \\
2                                                         & Teacher w/o NQG                                                               & 40.11          & 25.51           & 50.16           & 6.96          \\
3                                                         & Teacher w/o IQA                                                               & 39.54          & 24.87           & 49.45           & 6.54          \\
\rowcolor[HTML]{effafc}
4                                                         & Student (full)                                                                & 41.03          & 27.11           & 51.34           & 6.97          \\
5                                                         & Student w/o NTD                                                               & 37.54          & 25.04           & 48.26           & 5.54          \\
6                                                         & Student w/o Policy KL                                                         & 40.07          & 26.73           & 50.81           & 6.14          \\
7                                                         & Student w/o distillation                                                         & 29.41          & 19.89           & 39.91           & 4.01          \\
8                                                         & Student w/ uniform weights                                                    & 39.41          & 26.41           & 49.14           & 6.34          \\ \textcolor{black}{9}                                                         & \textcolor{black}{Attn.+Feat.+Logit KD~\cite{wang2024magic}}                                                    & \textcolor{black}{40.01}          & \textcolor{black}{26.59}           & \textcolor{black}{49.51}           & \textcolor{black}{6.39}          \\ \hline
\end{tabular}
}
\label{tab:ablate_core_soon}
\end{table}

\begin{figure*}[!t]
\centering
\includegraphics[width=\linewidth]{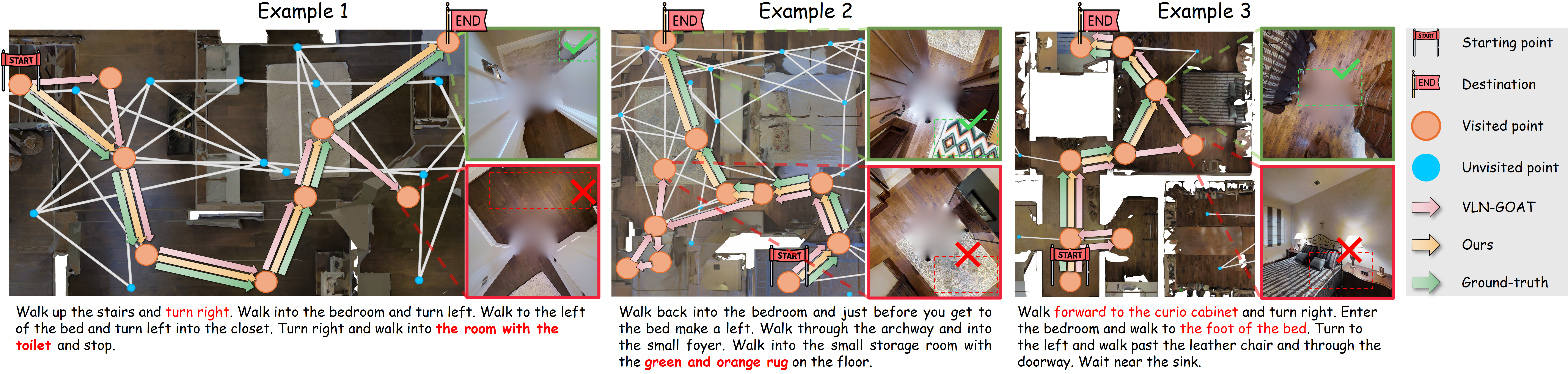}
\caption{\textcolor{black}{Qualitative trajectory comparison on R2R. 
The top-down graphs show navigation trajectories, the right panels show final viewpoints, and the corresponding instruction is provided below each episode. 
Yellow, green, and pink trajectories denote our student, the ground truth, and GOAT~\cite{wang2024causal}, respectively. 
Green boxes with check marks indicate correct grounding by our student, whereas red boxes indicate incorrect grounding by GOAT. 
Our student follows the instructions more faithfully, avoids detours, and reaches the correct destinations.}}
\label{fig:vis1}
\end{figure*}

\begin{table*}[!t]   
\centering
% \caption{\textbf{Architecture scaling ablation on R2R (Test Unseen).}
% We vary the depth of the text and panoramic encoders and the number of attention blocks in NQG and IQA (denoted as $N(\text{text})$, $N(\text{pano})$, $N(\text{NQG})$, and $N(\text{IQA})$), and report the resulting accuracy--efficiency trade-offs in terms of SR/SPL, parameters, and GFLOPs.}
\caption{\textcolor{black}{Architecture-scaling ablation on R2R test-unseen. We vary the depths of the text encoder, panoramic encoder, Navigable Query Generator, and Instruction--Query Aligner. Performance and efficiency are evaluated using SR, SPL, parameter count, and GFLOPs.}}
\resizebox{\linewidth}{!}{
\begin{tabular}{c|cccc|cc|cc}
\hline
\rowcolor[HTML]{F2F2F2} 
\cellcolor[HTML]{F2F2F2}                                  & \cellcolor[HTML]{F2F2F2}                                   & \cellcolor[HTML]{F2F2F2}                                   & \cellcolor[HTML]{F2F2F2}                                  & \cellcolor[HTML]{F2F2F2}                                  & \multicolumn{2}{c|}{\cellcolor[HTML]{F2F2F2}\textbf{Test Unseen}} & \cellcolor[HTML]{F2F2F2}                                      & \cellcolor[HTML]{F2F2F2}                                   \\
\rowcolor[HTML]{F2F2F2} 
\multirow{-2}{*}{\cellcolor[HTML]{F2F2F2}\textbf{Option}} & \multirow{-2}{*}{\cellcolor[HTML]{F2F2F2}\textbf{N(text)}} & \multirow{-2}{*}{\cellcolor[HTML]{F2F2F2}\textbf{N(pano)}} & \multirow{-2}{*}{\cellcolor[HTML]{F2F2F2}\textbf{N(NQG)}} & \multirow{-2}{*}{\cellcolor[HTML]{F2F2F2}\textbf{N(IQA)}} & \textbf{SR↑}                    & \textbf{SPL↑}                   & \multirow{-2}{*}{\cellcolor[HTML]{F2F2F2}\textbf{Param (M)↓}} & \multirow{-2}{*}{\cellcolor[HTML]{F2F2F2}\textbf{GFLOPs↓}} \\ \hline
\rowcolor[HTML]{effafc}
1 (Teacher)                                                        & 6                                                          & 2                                                          & 4                                                         & 3                                                         & 80.43                           & 71.98                           & 150.78                                                        & 45.56                                                      \\
2                                                         & 6                                                          & 2                                                          & 2                                                         & 1                                                         & 80.01                           & 71.21                           & 129.62                                                        & 38.87                                                      \\
3                                                         & 3                                                          & 1                                                          & 4                                                         & 3                                                         & 79.81                           & 70.12                           & 52.81                                                         & 15.57                                                      \\
4                                                         & 3                                                          & 1                                                          & 2                                                         & 1                                                         & 79.32                           & 69.94                           & 31.65                                                         & 8.06                                                       \\
\rowcolor[HTML]{effafc}
5 (Student)                                                         & 1                                                          & 1                                                          & 1                                                         & 1                                                         & 79.12                           & 69.54                           & 9.56                                                          & 1.92                                                       \\ \hline
\end{tabular}
}
\label{tab:ablate_depth}
\end{table*}

\begin{table}[t]   
\centering
% \caption{\textbf{Distillation target ablation on R2R (Test Unseen).}
% We compare distilling only global navigable queries, only local navigable queries, and distilling both branches (full) to quantify the complementary roles of global topological evidence and local one-step candidate evidence.}
\caption{\textcolor{black}{Distillation-target ablation on R2R test-unseen, comparing global-only, local-only, and joint global/local navigable-query distillation.}}
\resizebox{\linewidth}{!}{
\begin{tabular}{c|c|cccc}
\hline
\rowcolor[HTML]{F2F2F2} 
\cellcolor[HTML]{F2F2F2}                                  & \cellcolor[HTML]{F2F2F2}                                 & \multicolumn{4}{c}{\cellcolor[HTML]{F2F2F2}\textbf{Test   Unseen}} \\
\rowcolor[HTML]{F2F2F2} 
\multirow{-2}{*}{\cellcolor[HTML]{F2F2F2}\textbf{Option}} & \multirow{-2}{*}{\cellcolor[HTML]{F2F2F2}\textbf{Model}} & \textbf{SR↑}   & \textbf{SPL↑}   & \textbf{OSR↑}   & \textbf{NE↓}  \\ \hline
1                                                         & Global Only                                              & 75.41          & 65.13           & 80.81           & 3.05          \\
2                                                         & Local Only                                               & 74.22          & 64.34           & 80.23           & 3.11          \\
\rowcolor[HTML]{effafc}
3                                                         & Both (Full)                                              & 79.12          & 69.54           & 83.21           & 2.57          \\ \hline
\end{tabular}
}
\label{tab:ablate_distill_target_r2r}
\end{table}

\begin{table}[!t]   
\centering
% \caption{\textbf{Distillation target ablation on SOON (Test Unseen).}
% We compare distilling only global navigable queries, only local navigable queries, and distilling both branches (full) to quantify the complementary roles of global topological evidence and local one-step candidate evidence.}
\caption{\textcolor{black}{Distillation-target ablation on SOON test-unseen, comparing global-only, local-only, and joint global/local navigable-query distillation.}}
\label{tab:ablate_distill_target}
\resizebox{\linewidth}{!}{
\begin{tabular}{c|c|cccc}
\hline
\rowcolor[HTML]{F2F2F2} 
\cellcolor[HTML]{F2F2F2}                                  & \cellcolor[HTML]{F2F2F2}                                 & \multicolumn{4}{c}{\cellcolor[HTML]{F2F2F2}\textbf{Test   Unseen}} \\
\rowcolor[HTML]{F2F2F2} 
\multirow{-2}{*}{\cellcolor[HTML]{F2F2F2}\textbf{Option}} & \multirow{-2}{*}{\cellcolor[HTML]{F2F2F2}\textbf{Model}} & \textbf{SR↑}   & \textbf{SPL↑}   & \textbf{OSR↑}   & \textbf{RGSPL↑}  \\ \hline
1                                                         & Global Only                                              & 37.94          & 23.37           & 47.64           & 4.94         \\
2                                                         & Local Only                                               & 38.61          & 24.51           & 49.01           & 5.11          \\
\rowcolor[HTML]{effafc}
3                                                         & Both (Full)                                              & 41.03          & 27.11           & 51.34           & 6.97          \\ \hline
\end{tabular}
}
\end{table}

\subsubsection{Ablation on core modules and distillation objectives.}
Tab.~\ref{tab:ablate_core1} and Tab.~\ref{tab:ablate_core_soon} validate the contribution of each component on R2R and SOON, respectively.
On the teacher side, removing either NQG (Option~2) or IQA (Option~3) consistently degrades performance on both benchmarks, confirming that explicit navigable query extraction (Query) and instruction--query grounding (Align) are both essential for building a strong teacher.
In particular, the degradation appears not only on fine-grained navigation metrics in R2R but also on goal-oriented navigation and grounding metrics in SOON, showing that the proposed query-based evidence selection benefits both instruction following and object-oriented navigation.

On the student side, removing NTD (Option~5) leads to a clear performance drop, indicating that transferring action-sufficient navigable queries is crucial for compact models.
Removing policy KL (Option~6) also hurts performance, which suggests that behavior-level decision supervision complements query-level representation distillation during fine-tuning.
When both NTD and policy KL are removed (Option~7), the student suffers the largest degradation, especially on SOON, demonstrating that supervised training alone is insufficient for a lightweight student to recover the teacher's evidence-selection and decision-making abilities.
Finally, replacing the token-adaptive weighting strategy with uniform weights (Option~8) consistently underperforms the full student, verifying that emphasizing poorly aligned query tokens enables more effective knowledge transfer than naïve equal-weight matching.

\textcolor{black}{
We further compare NTD with an attention-, feature-, and logit-level KD baseline following the target types used in MAGIC, while keeping the same frozen teacher, compact student, task objectives, and training schedule. 
On R2R, the full QAD student improves SR/SPL/OSR by 0.81/3.00/1.69 points over this generic KD baseline and reduces NE by 0.12. 
On SOON, it improves SR/SPL/OSR/RGSPL by 1.02/0.52/1.83/0.58 points.
}

\textcolor{black}{
This controlled comparison indicates that directly transferring the structured navigable-query interface is more effective than matching generic intermediate attentions and features. 
Moreover, removing NTD while retaining policy KL reduces SPL from 69.54 to 61.44 on R2R and from 27.11 to 25.04 on SOON, showing that standard policy distillation alone cannot account for the performance gain. 
Finally, uniform query matching remains inferior to NTD, confirming that both the navigation-specific query target and slot-level adaptive weighting contribute to effective knowledge transfer.
}

\begin{figure*}[t]
\centering
\includegraphics[width=\linewidth]{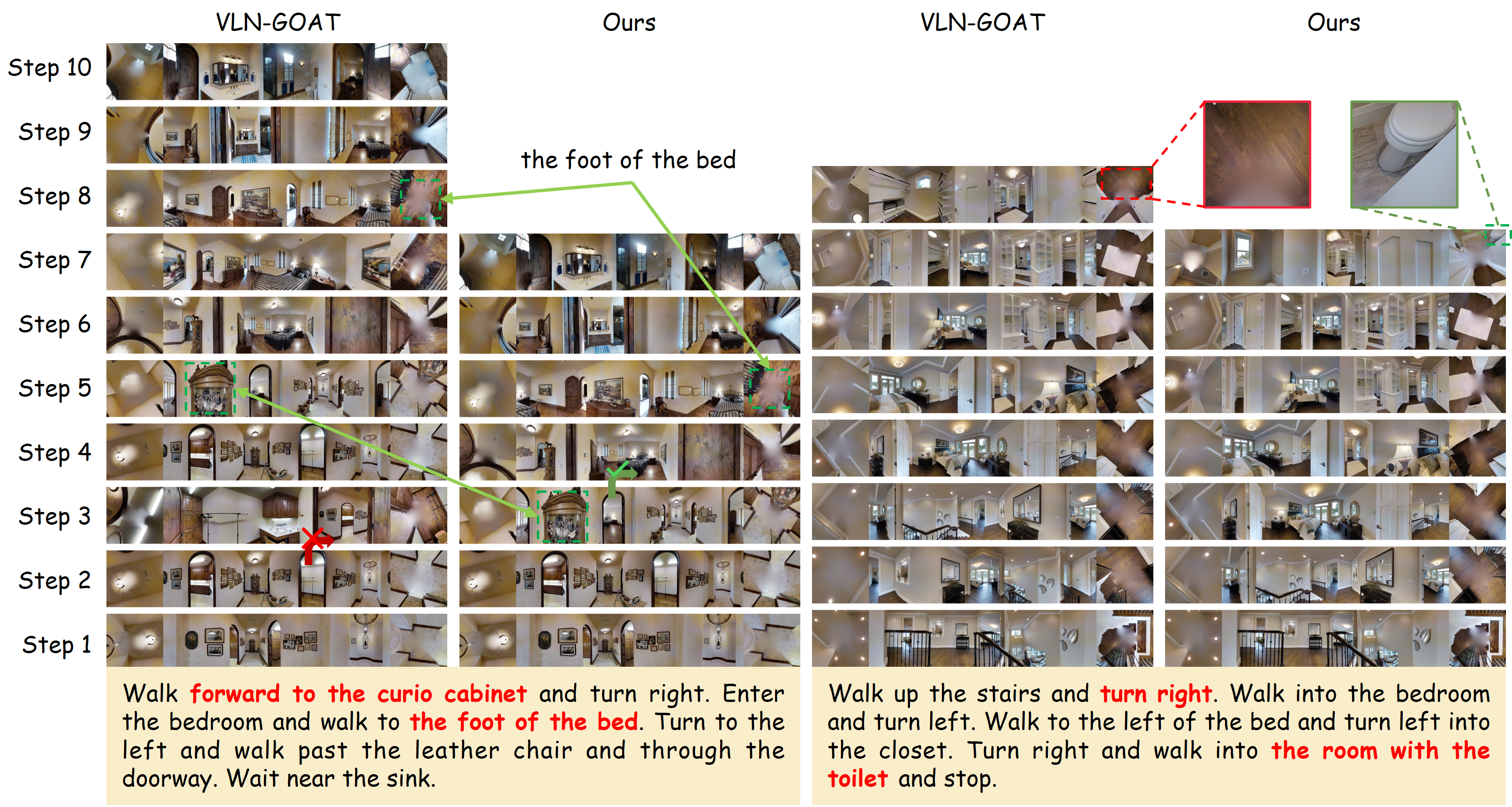}
\caption{\textcolor{black}{Step-wise visualization of the navigation process. 
Each row shows the panoramic observation at one navigation step, and the input instruction is provided at the bottom with key cues highlighted in red. 
In the left example, the green check and red cross denote correct and incorrect turning decisions, respectively. 
In the right example, the green and red boxes indicate correctly and incorrectly grounded visual evidence, respectively. 
Our student follows the highlighted cues more accurately and reaches the destination with a shorter trajectory than GOAT~\cite{wang2024causal}.}}
\label{fig:vis2}
\end{figure*}

\begin{figure*}[t]
\centering
\includegraphics[width=\textwidth]{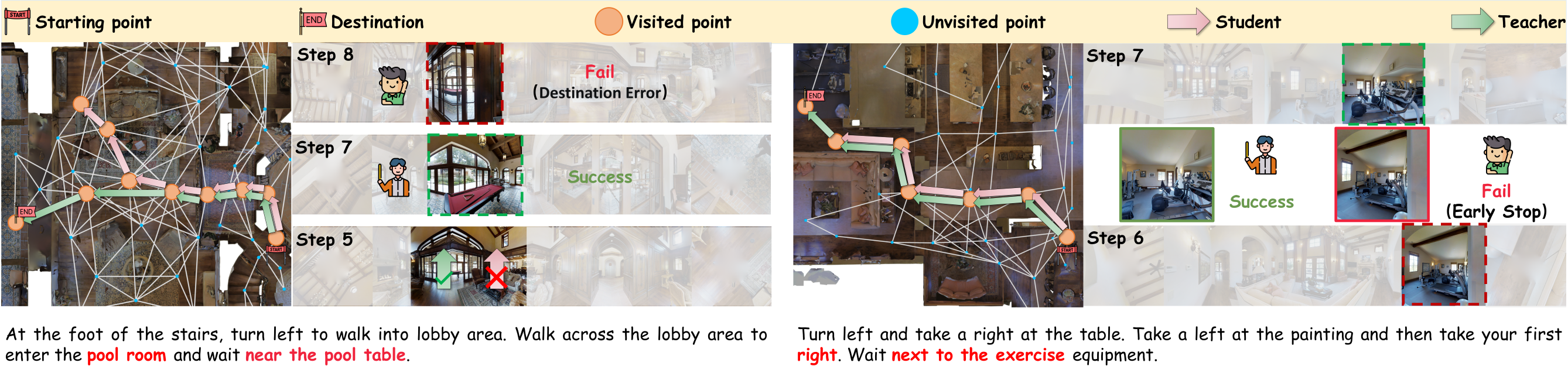}
\caption{
\textcolor{black}{Teacher--student failure cases. The teacher succeeds while the compact student
fails. The left case shows a destination-room grounding error: the student fails
to associate ``pool room / pool table'' with the correct visual evidence. The
right case shows a premature STOP error: the student stops once exercise
equipment becomes visible, while the teacher moves to the correct viewpoint
specified by ``next to the exercise equipment.''
}
}
\label{fig:failure_cases}
\end{figure*}

\begin{figure}[t]
\centering
\includegraphics[width=\linewidth]{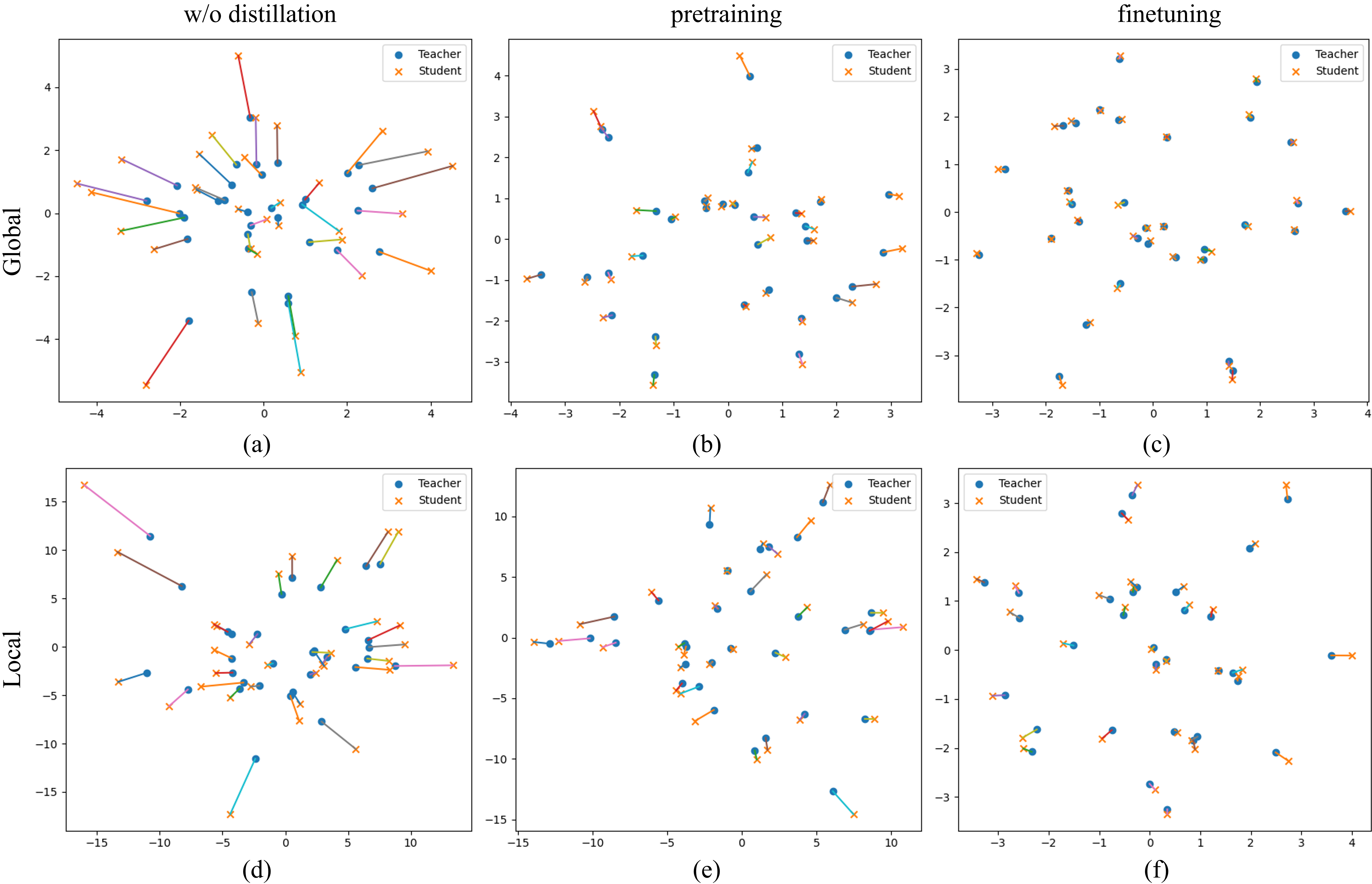}
\caption{\textcolor{black}{PCA visualization of teacher--student query alignment.
Global and local navigable-query slots (\(K{=}32, D{=}768\)) are projected into a shared 2D PCA space across three training stages.
Blue circles and orange crosses denote teacher and student slots, respectively, and each line connects a same-index pair.
Shorter links indicate stronger slot-wise alignment.}}
\label{fig:pca_query_alignment}
\end{figure}

\subsubsection{Ablation on distillation targets (global vs. local queries).}
Tab.~\ref{tab:ablate_distill_target_r2r} and Tab.~\ref{tab:ablate_distill_target} compare different distillation targets on R2R and SOON.
Distilling only global or only local navigable queries causes clear degradation, while distilling both branches consistently achieves the best performance.
This indicates that the two branches encode complementary navigation evidence: global queries capture long-horizon topological intent, whereas local queries preserve fine-grained actionability for immediate candidate selection.
The trend holds for both fine-grained instruction following on R2R and goal-oriented navigation on SOON, confirming the necessity of jointly transferring global and local navigable evidence to the compact student.

\subsubsection{Ablation on architecture scaling.} \textcolor{black}{Tab.}~\ref{tab:ablate_depth} shows that most of the computation and parameter reduction comes from shrinking the encoder depth and cross-modal reasoning depth, while the performance drop is relatively mild until we reach the most compact setting.
Starting from the teacher (Option~1), reducing only the cross-modal depth (Option~2) yields a noticeable reduction in Params/GFLOPs with minimal SR/SPL degradation, indicating that our query bottleneck enables a shallower reasoning stack without collapsing navigation accuracy.
Reducing only the encoder depth (Option~3) brings a much larger efficiency gain, but introduces a clearer accuracy drop, suggesting that representation capacity in text/panoramic encoders is still important.
Combining both reductions (Option~4) further improves efficiency with a controlled performance loss.
Finally, the compact student (Option~5) achieves a favorable accuracy--efficiency frontier, approaching teacher-level SR/SPL with orders-of-magnitude lower GFLOPs and parameters.

\subsubsection{Ablation on the number of navigable query slots.}
\textcolor{black}{
We further study the influence of the per-branch query-slot number $K_b$ in Fig.~\ref{fig:ablation1}(b), where $K_L=K_G=K_b$ and $K_{\rm tot}=2K_b$. All remaining architectural and training settings are kept unchanged. The query slots summarize navigation cues from the local and global branches. When $K_b$ is small, only a few slots are available to represent multiple cues, such as candidate directions, landmarks, navigation progress, or backtracking information. Different cues may therefore be mixed in the same slots, causing useful information to be lost. This explains the lower SR and SPL at small values of $K_b$.
}

\textcolor{black}{
As shown in Fig.~\ref{fig:ablation1}(b), increasing $K_b$ from 8 to 32 steadily improves both SR and SPL because more slots allow the model to retain more navigation information. Beyond 32, SR changes only modestly, while SPL no longer improves. This suggests that most useful cues have already been captured and that additional slots mainly represent overlapping information. Increasing the number of slots also requires more computation because query extraction is performed for every slot and self-attention compares all query pairs. We therefore set $K_L=K_G=32$ ($K_{\rm tot}=64$), which gives the best SPL and a strong SR with fewer query slots. The best value may vary for other architectures or datasets.
}

\subsubsection{Ablation on KL weighting.}
The policy distillation term $\mathcal{L}_{\pi}$ is introduced only during fine-tuning, where the objective involves task-specific action prediction.
To balance representation-level query transfer and behavior-level supervision, we optimize
$\mathcal{L}_{D}=\alpha\,\mathcal{L}_{\mathrm{NTD}}+(1-\alpha)\,\mathcal{L}_{\pi}$,
and conduct a sweep over $\alpha$ on both R2R and REVERIE (\textcolor{black}{Fig.}~\ref{fig:alpha_ablation}).
We observe that performance is stable within a reasonable range of $\alpha$, and a moderate value achieves the strongest results.
When $\alpha$ is too large, the student relies mainly on query matching and under-utilizes policy guidance; when $\alpha$ is too small, the student is overly constrained by the teacher’s action distribution, which can weaken the benefit of token-level intent transfer.

\subsection{Visualization Analysis}
\subsubsection{Trajectory-level comparison.}
\textcolor{black}{Fig.}~\ref{fig:vis1} shows three episodes with trajectories of GOAT, ours, and the ground-truth path.
Our student follows instruction cues more reliably, avoiding detours and reducing premature stops at visually plausible but semantically incorrect locations, leading to shorter and more efficient routes—especially when instructions involve fine-grained turns or landmark cues.

\subsubsection{Step-wise grounding.}
\textcolor{black}{Fig.}~\ref{fig:vis2} examines step-by-step panoramas.
Our student grounds key phrases (e.g., \emph{``the foot of the bed''}, \emph{``the room with the toilet''}) to the correct visual evidence across steps, whereas GOAT is more easily distracted by ambiguous textures or clutter, which can trigger wrong transitions.
These qualitative results support our quantitative improvements, suggesting that stronger phrase-level grounding yields more accurate decisions and shorter trajectories even with a compact student.

\begin{figure}[t]
\centering
\includegraphics[width=\linewidth]{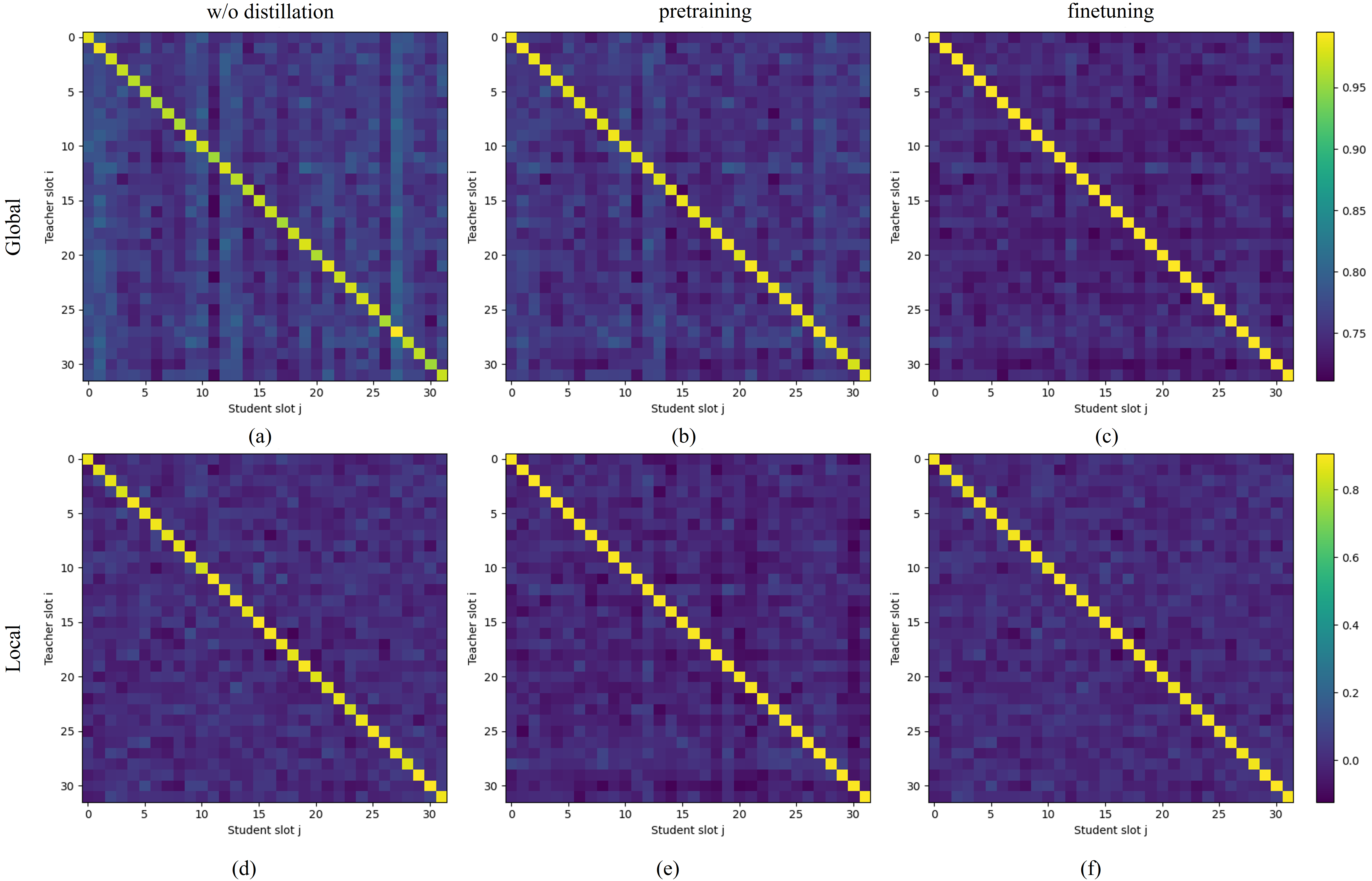}
\caption{\textcolor{black}{{Cosine similarity matrices for teacher--student query slots.}
Each matrix $\mathbf{S}\in\mathbb{R}^{K\times K}$ ($K{=}32$) measures
$\mathbf{S}_{ij}=\cos(\mathbf{q}^{T}_{i},\mathbf{q}^{S}_{j})$.
All panels share the same color scale.
Stronger diagonal responses indicate better same-index alignment, whereas off-diagonal responses suggest slot permutation or mixing.
Distillation progressively sharpens the diagonal structure for both global and local queries.}}
\label{fig:sim_query_alignment}
\end{figure}

\subsubsection{Teacher--student failure analysis.}
\textcolor{black}{
To further diagnose what navigation skills are still weakened after distillation, we analyze representative episodes where the QAD teacher succeeds, but the compact student fails. 
As shown in Fig.~\ref{fig:failure_cases}, the student errors mainly come from two types of fine-grained decision failures.
}

\textcolor{black}{
In the first example, the instruction requires the agent to ``enter the pool room'' and ``wait near the pool table.'' 
The teacher correctly follows the route, enters the pool-room area, and stops near the pool table. 
In contrast, the student deviates toward a visually plausible but semantically incorrect destination.
This indicates that the compact student has weaker destination-room and landmark grounding, especially when the target phrase requires recognizing a room type from its contained object cues, e.g., associating ``pool room'' with the visual evidence of a pool table.
}

\textcolor{black}{
In the second example, the instruction asks the agent to ``wait next to the
exercise equipment.'' The student already observes exercise equipment from an
adjacent viewpoint and therefore triggers the STOP action prematurely. However,
the teacher continues one more step and stops at the correct viewpoint next to
the exercise equipment. This suggests that the student has learned coarse
object-presence cues, but its fine-grained STOP decision boundary and spatial
relation grounding, e.g., ``next to,'' are still less precise than those of the teacher.
}

\textcolor{black}{
These examples indicate that QAD transfers most navigation evidence from the
teacher to the student, but the remaining performance gap mainly arises from
fine-grained visual-language grounding and precise stop/action calibration.
This is consistent with our ablation results: removing NTD causes a clear
performance drop, and distilling both global and local queries outperforms
distilling either branch alone, suggesting that both long-horizon topological
evidence and local action-feasible evidence are necessary for preserving the
teacher's navigation skills.
}

\subsubsection{Query alignment analysis}
To verify that our query-level distillation effectively transfers the teacher's \emph{query representations} to the student, we visualize the alignment of both \textbf{global} and \textbf{local} navigable queries across training stages.
In \textcolor{black}{Fig.}~\ref{fig:pca_query_alignment}, the \textbf{first row (a--c)} shows \textbf{global} queries and the \textbf{second row (d--f)} shows \textbf{local} queries,
where each teacher slot is connected to the student slot with the same index \(k\).

Under \textbf{ad} (independent training; \textcolor{black}{Fig.}~\ref{fig:pca_query_alignment} a,d), the teacher and student queries are widely separated in the projected space,
and the index-wise links are generally long, suggesting that the student fails to capture the teacher's navigable-query representations when trained without distillation. With \textbf{be} (pretraining distillation; \textcolor{black}{Fig.}~\ref{fig:pca_query_alignment} b,e), the distances between paired teacher--student slots shrink noticeably,
indicating that the student begins to absorb the teacher's query representation structure through query-level supervision.
After \textbf{cf} (additional finetuning distillation; \textcolor{black}{Fig.}~\ref{fig:pca_query_alignment} c,f), the paired slots become the closest,
showing the strongest teacher--student representation alignment.

This observation is consistent with the cosine similarity matrices in \textcolor{black}{Fig.}~\ref{fig:sim_query_alignment}.
From \textbf{ad} to \textbf{be} and further to \textbf{cf}, diagonal similarities (Teacher[\(k\)] vs. Student[\(k\)]) become progressively higher for both
global (a--c) and local (d--f) queries, while non-diagonal similarities remain comparatively lower.
Overall, these qualitative results support that our distillation progressively improves the student's ability to capture the teacher's navigable-query representations,
thereby transferring compact \emph{where-to-attend} evidence into the student.

\subsection{Discussion}
\textcolor{black}{
Our Jetson Nano experiments provide an initial evaluation of QAD's inference efficiency on a resource-constrained edge platform. 
The student achieves 51.6 steps/s under FP32 inference, demonstrating its computational feasibility on resource-constrained hardware. 
Nevertheless, simulator-to-real transfer remains challenging. 
Simulator benchmarks rely on pre-extracted panoramic features and graph-based navigation in relatively stable scenes, whereas real-world environments are affected by illumination changes, occlusions, motion blur, and sensor noise. 
These factors may impair visual grounding and cause navigation errors to accumulate over time. 
To narrow this gap, future work will focus on two aspects. First, we will train QAD under changing illumination and random obstacles. Second, we will integrate QAD with lightweight onboard perception and control modules before evaluating the complete system on physical robots in dynamic indoor environments.
}

\section{Conclusion}
We presented \textbf{Query, Align, and Distill (QAD)}, a navigation-aware teacher--student framework for efficient Vision-and-Language Navigation.
We achieve efficient VLN in two steps: \textbf{(1) build a strong teacher} that makes evidence selection explicit and compressible by extracting action-sufficient \emph{global/local} navigable evidence with a small set of learnable queries and progressively aligning these evidence tokens with instruction semantics for policy prediction; and \textbf{(2) distill the capability to a compact student} by transferring both \emph{where to attend} via Navigation-aware Token-adaptive Distillation and \emph{what to do} via policy distillation.
Extensive experiments across fine-grained and goal-oriented VLN benchmarks show that the student retains near-teacher navigation accuracy while substantially reducing computation and parameters, and delivers tangible runtime gains on edge devices.

\section*{Acknowledgments}
This work was supported by the Young Scientists Fund of NSFC (Grant No. 62406035), and Start-up Funding of Beijing University of Posts and Telecommunications (Grant No.510224072).

\bibliographystyle{unsrt}
\bibliography{ref}

\newpage
\vspace{-70mm}
\begin{IEEEbiography}[{\includegraphics[width=1in,height=1.25in,clip,keepaspectratio]{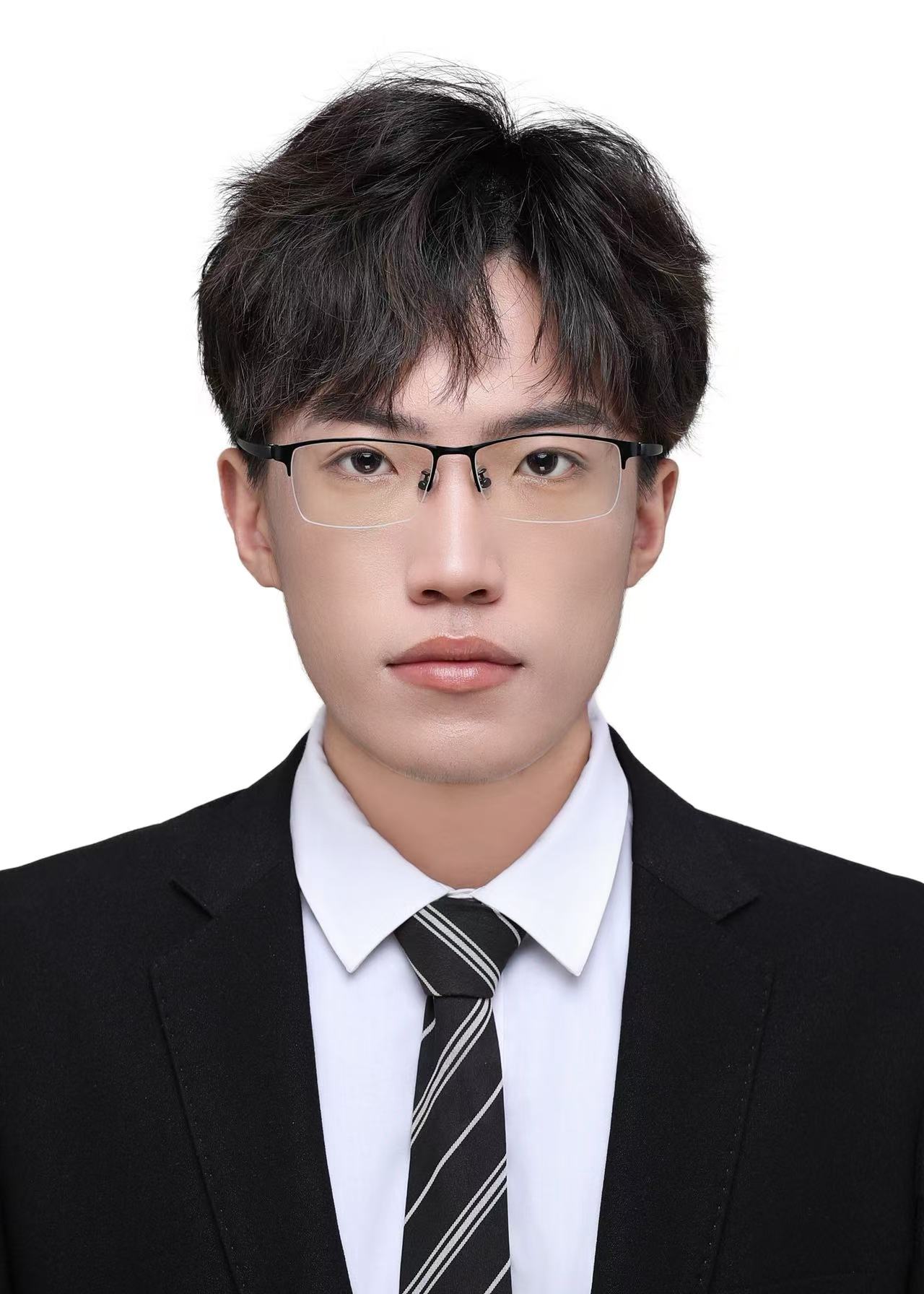}}]{Zhihao Chen} received the B.S. degree in the  Department of Software Engineering, School of Computer, Beijing Information Science and Technology University (BISTU), Beijing, China, in 2025. He is currently pursuing the M.S. degree in the School of Intelligent Engineering and Automation, Beijing University of Posts and Telecommunications (BUPT). His research interests focus on underwater image enhancement and person re-identification.
\end{IEEEbiography}
\vspace{-70mm}
\begin{IEEEbiography}[{\includegraphics[width=1in,height=1.25in,clip,keepaspectratio]{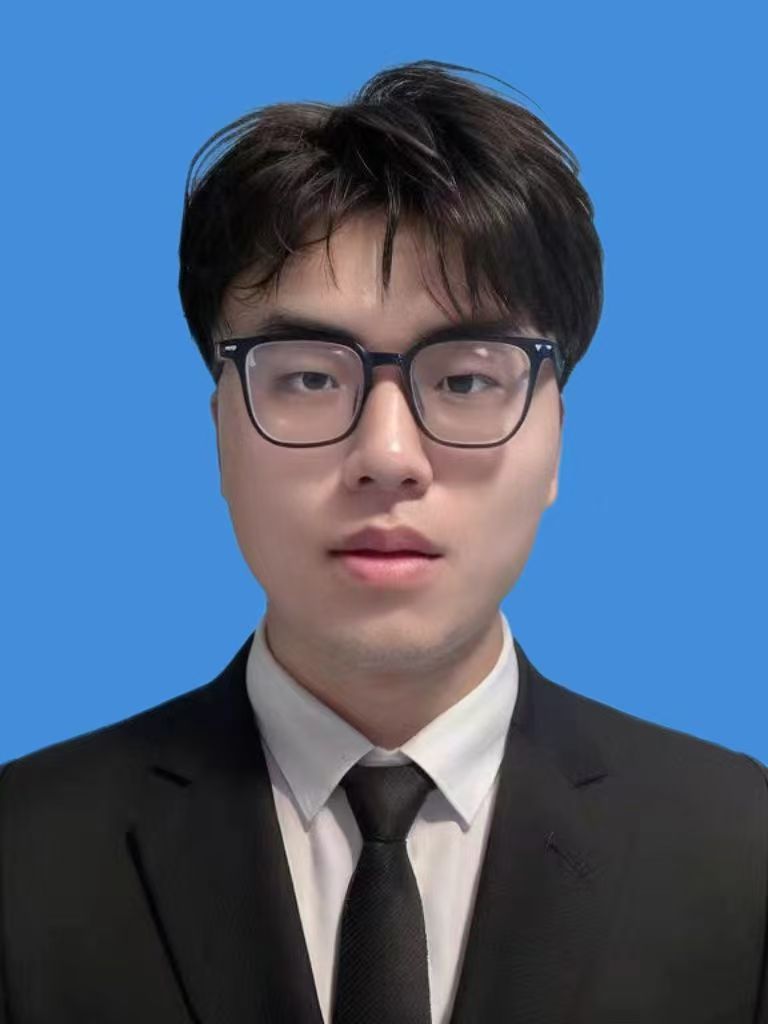}}]{Yiyuan Ge} received the B.S. degree in the National Demonstration Software College of Jilin University (JLU), Changchun, China, in 2022. He obtained the M.S. degree in the School of Instrument Science and Opto-Electronics Engineering, Beijing Information Science and Technology University (BISTU), in 2025. He is currently pursuing a Ph.D. degree in the School of Electronic and Information Science, South China University of Technology (SCUT). His research interests focus on low-level vision and person re-identification.
\end{IEEEbiography}
\vspace{-70mm}
\begin{IEEEbiography}[{\includegraphics[width=1in,height=1.25in,clip,keepaspectratio]{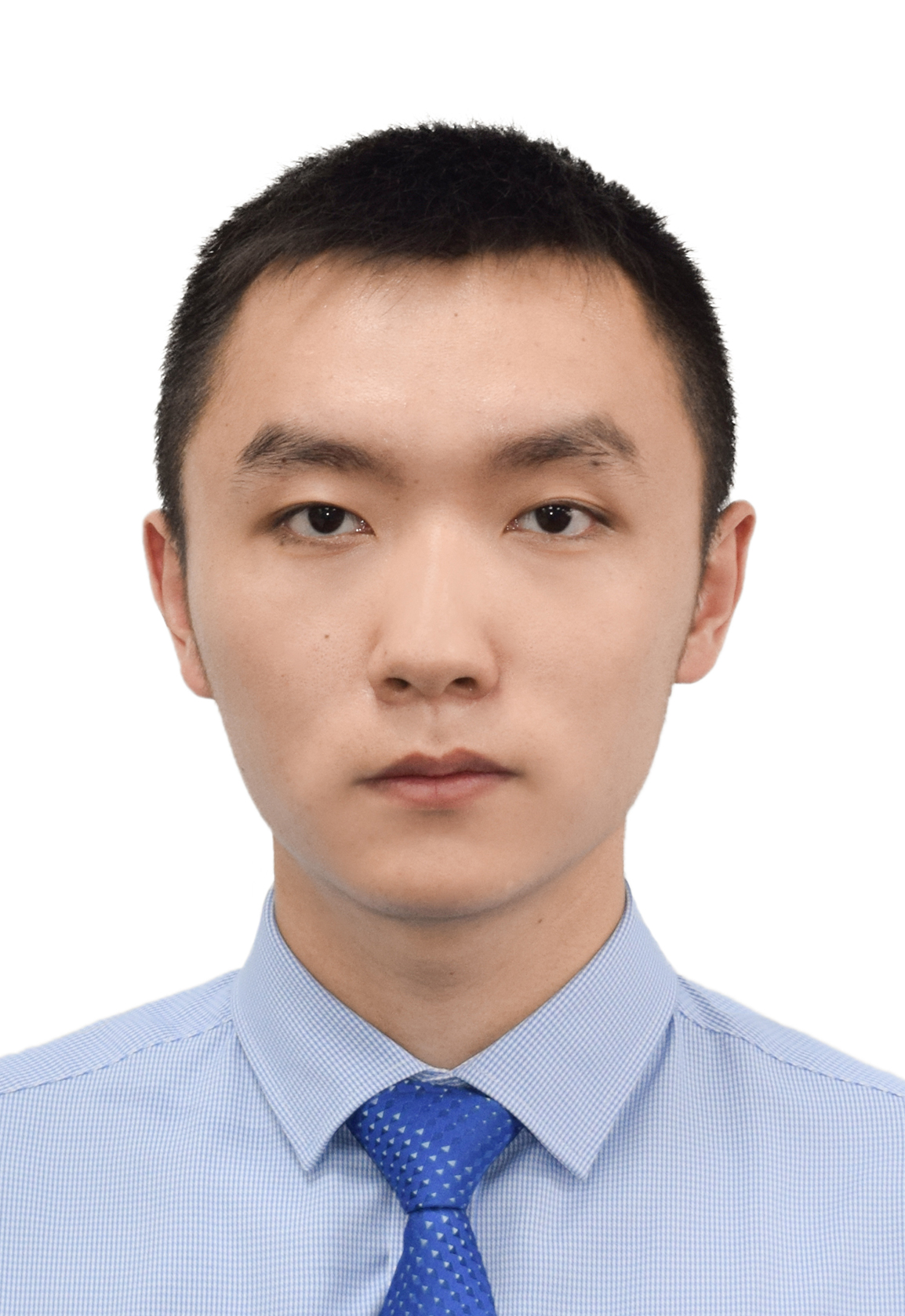}}]{Ziyang Wang} is a Research Fellow at The Alan Turing Institute, UK. Prior to this, he completed DPhil in Computer Science at the University of Oxford in 2024, MRes at Imperial College London in 2018, and BEng in Automation at Xi’an Jiaotong University in 2017. His research interests include Computer Vision, Data-Centric Engineering, Healthcare AI, and Robotics.  
\end{IEEEbiography}
\vspace{-70mm}
\begin{IEEEbiography}[{\includegraphics[width=1in,height=1.25in,clip,keepaspectratio]{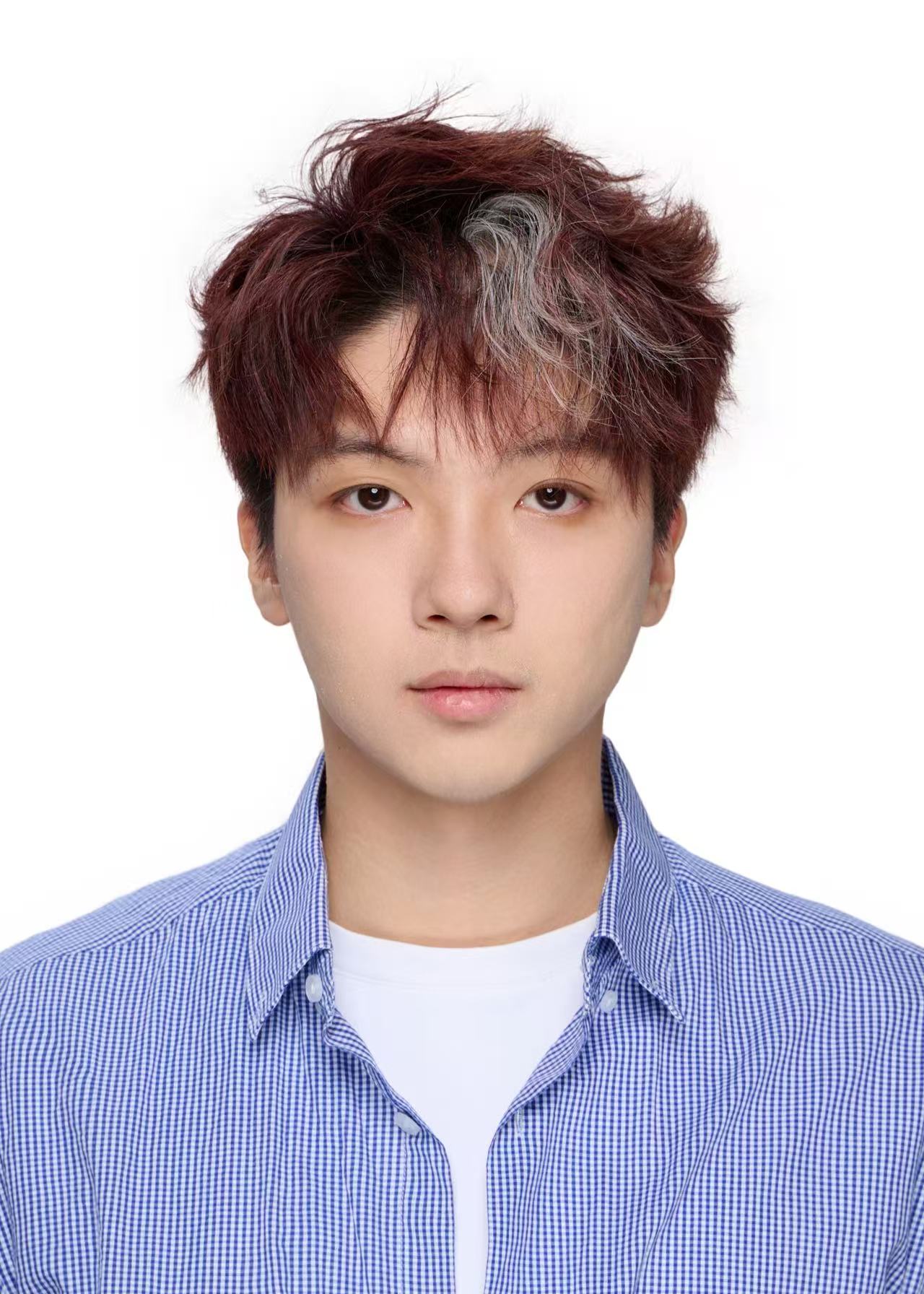}}]{Pu Cao}
received his bachelor’s degree from University of Science and Technology Beijing (USTB), Beijing, China, in 2022, and is currently a Ph.D. candidate at the School of Intelligent Engineering and Automation, Beijing University of Posts and Telecommunications (BUPT), since 2022. His research interests include multimodal understanding and generation, especially focusing on multimodal large-language models and diffusion models. \end{IEEEbiography}
\vspace{-70mm}
\begin{IEEEbiography}[{\includegraphics[width=1in,height=1.25in,clip,keepaspectratio]{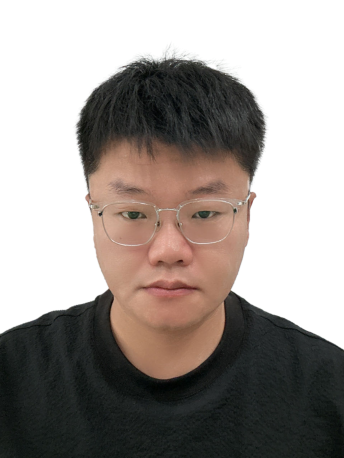}}]{Lu Yang} is currently an associate professor in the Beijing University of Posts and Telecommunications (BUPT), School of Intelligent Engineering and Automation, China. He received his Ph.D. degree from the BUPT in 2021. He has been involved in research work with the Pattern Recognition and Intelligent Vision Laboratory (PRIV), since 2012. His current research interests include the fields of Human-Centric Al and GenAI. \end{IEEEbiography}

\end{document}